\documentclass{article}

\usepackage{arxiv}

\usepackage[utf8]{inputenc} 
\usepackage[T1]{fontenc}    
\usepackage{hyperref}       
\usepackage{url}            
\usepackage{booktabs}       
\usepackage{amsfonts}       
\usepackage{nicefrac}       
\usepackage{microtype}      
\usepackage{lipsum}

\usepackage{pdflscape}
\usepackage{adjustbox}
\usepackage{rotating}
\usepackage{capt-of}

\usepackage{booktabs} 
\usepackage{graphicx}
\usepackage[vlined,linesnumbered,ruled,titlenotnumbered]{algorithm2e}
\usepackage{mathtools}
\usepackage{hyperref}
\usepackage{balance}
\usepackage{xspace}
\usepackage{comment}
\usepackage{algpseudocode,booktabs,amsmath}
\usepackage{multirow, rotating}
\usepackage{mwe}
\usepackage{color}
\usepackage{blindtext}
\usepackage{amssymb}
\usepackage{amsmath}
\usepackage{amsthm}
\usepackage{mathtools}

\usepackage{subcaption}
\usepackage{rotating}

\usepackage{hyperref}
\usepackage{balance}
\usepackage{tikz}
\usetikzlibrary{shapes,decorations,patterns,positioning}
\usetikzlibrary{plotmarks}
\usepackage{url}
\usepackage{multirow}
\usepackage{algorithm2e}

\newtheorem{theorem}             {Theorem}

\newcommand{\EA}{\mbox{\em{MOEA}}\xspace}
\newcommand{\EAd}{\mbox{\em{MOEA$_D$}}\xspace}

\DeclareSymbolFont{symbolsC}{U}{pxsyc}{m}{n}
\DeclareMathSymbol{\colonequals}{\mathrel}{symbolsC}{"42}

\title{Single- and Multi-Objective Evolutionary Algorithms for the Knapsack Problem with Dynamically Changing Constraints}

\author{
  Vahid Roostapour \\
  Optimisation and Logistics\\
  The University of Adelaide\\
  Adelaide, Australia \\
  \texttt{vahid.roostapour@gmail.com} \\
   \And
   Aneta Neumann \\
  Optimisation and Logistics\\
  The University of Adelaide\\
  Adelaide, Australia \\
  \texttt{aneta.neumann@adelaide.edu.au} \\
   \And
 Frank Neumann \\
  Optimisation and Logistics\\
  The University of Adelaide\\
  Adelaide, Australia \\
  \texttt{frank.neumann@adelaide.edu.au} \\
}

\begin{document}
\maketitle

\begin{abstract}
Evolutionary algorithms are bio-inspired algorithms that can easily adapt to changing environments.  Recent results in the area of runtime analysis have pointed out that algorithms such as the (1+1)~EA and Global SEMO can efficiently reoptimize linear functions under a dynamic uniform constraint.
Motivated by this study, we investigate single- and multi-objective baseline evolutionary algorithms for the classical knapsack problem where the capacity of the knapsack varies over time.

We establish different benchmark scenarios where the capacity changes every $\tau$ iterations according to a uniform or normal distribution. Our experimental investigations analyze the behavior of our algorithms in terms of the magnitude of changes determined by parameters of the chosen distribution, the frequency determined by $\tau$, and the class of knapsack instance under consideration. Our results show that the multi-objective approaches using a population that caters for dynamic changes have a clear advantage on many benchmarks scenarios when the frequency of changes is not too high. Furthermore, we demonstrate that the diversity mechanisms used in popular evolutionary multi-objective algorithms such as NSGA-II and SPEA2 do not necessarily result in better performance and even lead to inferior results compared to our simple multi-objective approaches.
\end{abstract}


\section{Introduction}
Evolutionary algorithms~\cite{DBLP:series/ncs/EibenS15} have been widely applied to a wide range of combinatorial optimization problems. They often provide good solutions to complex problems without a large design effort. Furthermore, evolutionary algorithms and other bio-inspired computing have been applied to many dynamic and stochastic problems~\cite{DCOPS,DBLP:journals/swevo/RakshitKD17,DBLP:conf/aaai/DoerrD0NS20, DBLP:conf/ppsn/NeumannN20} as they have the ability to easily adapt to changing environments.

Most studies for dynamic problems so far focus on dynamic fitness functions~\cite{DBLP:journals/swevo/NguyenYB12}. However, in real-world applications the optimization goal such as maximizing profit or minimizing costs often does not change. Instead, resources to achieve this goal change over time and influence the quality of solutions that can be obtained. 
In the context of continuous optimization, dynamically changing constraints have been investigated in~\cite{DCOPS,DBLP:conf/evoW/Ameca-AlducinHN18}.

Theoretical investigations for combinatorial optimization problems with dynamically changing constraints have recently been carried out on different types of problems~\cite{DBLP:journals/corr/abs-1806-08547}. Studies investigated the efficiency of algorithms in finding good quality solutions from scratch when criteria change dynamically during the optimization process. Furthermore, studies considered the time to adapt good solutions after a dynamic change happens. \cite{DBLP:conf/gecco/PourhassanGN15} and ~\cite{DBLP:conf/gecco/Bossek0PS19} analyzed the performance of simple algorithms on vertex cover and graph coloring problems, respectively, in dynamic environment, where a dynamic change might add or remove an edge to or from the graph.~\cite{DBLP:conf/aaai/RoostapourN0019} studied the class of submodular functions where the constraint bound changes dynamically. They proved that a simple multi-objective evolutionary approach called POMC efficiently guarantees the same worst-case approximation ratio in the considered dynamic environment as the worst-case approximation ratio of classical greedy algorithms in the static environment. In another general study,~\cite{DBLP:journals/algorithmica/ShiSFKN19} considered baseline evolutionary algorithms such as the (1+1)~EA and Global SEMO dealing with linear functions under dynamic uniform constraints. This study investigates the runtime of finding a new optimal solution after the constraint bound has changed by certain amount. They theoretically analysed the uniform constraint version since the problem is NP-hard under a linear constraint, which is equivalent to the knapsack problem.

The goal of this paper is to contribute to the research on how evolutionary algorithms can deal with dynamic constraints from an experimental perspective and therefore complement the beforehand mentioned theoretical studies.
A dynamic version of the multi-dimensional knapsack problem has been studied in the literature~\cite{DBLP:conf/gecco/BrankeSU05,10.1007/11732242_74,10.1007/978-3-642-01129-0_86}. Here the knapsack problem has several capacity constraints. In the dynamic setting described in \cite{DBLP:conf/gecco/BrankeSU05}, the profits, weights, and constraint bounds change simultaneously and in a multiplicative way according to predefined normal distributions. 
Our goal is to study changes to a single constraint bound systematically in order to provide insights that complement recent work in the area of runtime analysis. 
To experimentally investigate evolutionary algorithms for the knapsack problem with a dynamically changing capacity, we design a specific benchmark set. This benchmark set is built on classical static knapsack instances and varies the constraint bound over time.
The changes of the constraint bound occur randomly every $\tau$ iterations, where $\tau$ is a parameter determining the frequency of changes. The magnitude of a change is chosen from a uniform distribution in an interval $[-r, r]$, where $r$ determines the magnitude of changes. Furthermore, we examine changes according to the normal distribution $\mathcal{N}(0, \sigma^2)$ with mean $0$ and standard deviation $\sigma$. Here $\sigma$ is used to determine the magnitude of changes. Note that larger values of $\sigma$ make larger changes more likely.

The comparison between the algorithms considered is based on the offline errors. We compute the exact optimal solutions for each possible capacity by performing dynamic programming in a preprocessing phase. To calculate the total offline error, we consider the difference between the profit of the best achieved feasible solution in each iteration and the profit of the optimal solution. The total offline error illustrates the performance of the algorithms during the optimization process and demonstrates the quality of the feasible solution.

In the second part of the paper we consider advanced evolutionary multi-objective algorithms, where, in addition to the total offline error, we carry out comparisons with respect to the partial offline error. The partial offline error only considers the best feasible solution found by an algorithm exactly before the next dynamic change happens. This factor does not illustrate how fast the algorithm finds a good quality solution. In this way, instead of analyzing the performance during the optimization period, we study the algorithms based on their final results for the different changes that occurred.

The first part of our experimental analysis, which investigates the theoretical results in~\cite{DBLP:journals/algorithmica/ShiSFKN19}, corresponds to examining the uniform instances in which all the weights are one. We consider the performance of (1+1)~EA against two different versions of simple Multi-Objective Evolutionary Algorithms (\EA and \EAd, which only differ in their definition of dominance). Both are inspired by the Global SEMO algorithm which has frequently been used in successful Pareto optimization approaches for problems with various types of constraints~\cite{DBLP:journals/nc/NeumannW06,DBLP:journals/ec/FriedrichHHNW10,DBLP:journals/algorithmica/KratschN13,DBLP:journals/ec/FriedrichN15,DBLP:books/sp/ZhouYQ19}. 
The multi-objective approaches store infeasible solutions as part of the population in addition to feasible solutions. In contrast to Global SEMO which usually stores all trade-offs with respect to the given objective functions, we limit the algorithm to keep only solutions around the current constraint bound. The range of feasible and infeasible solutions stored in our multi-objective algorithms is set based on the anticipated change of the constraint bound. Our experimental analysis confirms the theoretical results given in \cite{DBLP:journals/algorithmica/ShiSFKN19} on our benchmarks with dynamic uniform constraints and shows that the multi-objective approaches outperform (1+1)~EA. 

Afterwards, we study the performance of the same baseline algorithms dealing with general version of the dynamic knapsack problem, in which the weights are not uniform.
For the general setting, we investigate different instances of the knapsack problem, such as instances with randomly chosen weights and profits, and instances with strongly correlated weights and profits. We study the behavior of the algorithms in various situations, where the frequency and the magnitude of changes are different. Our results show that the (1+1)~EA has an advantage over the multi-objective algorithms when the frequency of changes is high. In this case, the population slows down the adaptation to the changes. On the other hand, lower frequency of changes play in favor of the multi-objective approaches, in most of the cases. Exceptions occur in some situations where weights and profits are highly correlated or have similar values. 

This paper extends its conference version~\cite{DBLP:conf/ppsn/Roostapour0N18} by a theoretical runtime analysis, which compares the performance of the basic single- and multi-objective approaches for the knapsack problem with dynamically changing constraints, and an experimental investigation of the performance of NSGA-II~\cite{DBLP:journals/tec/DebAPM02} and SPEA2~\cite{zitzler2001spea2}, as representatives of advanced evolutionary multi-objective algorithms, in the dynamic environment.
We use jMetal package as the base of our implementations and modify the algorithms to perform on the dynamic KP~\cite{DBLP:journals/aes/DurilloN11}.
Each of the algorithms calculates a specific fitness value based on the non-dominance rank and position of each individual among the others in the objective space. We compare these algorithms with \EAd, as it outperforms the other baseline evolutionary algorithms, and investigate the performance of advanced techniques in NSGA-II and SPEA2 against the simple approach of \EAd. Note that such comparison requires some modification in the selection and fitness functions of NSGA-II and SPEA2 to make us able to compare their results with \EAd. Our experimental results illustrate that while NSGA-II and SPEA2 react faster to a dynamic change, \EAd can find better solutions if it has enough time before the next dynamic change. We also show that the techniques for producing well-distributed non-dominated solutions prevent these algorithms from improving their best feasible solution. We address this problem by presenting an additional elitism approach to be introduced into NSGA-II and SPEA2 and show its benefit on our benchmark set.

The outline of the paper is as follows. Section~\ref{sec:prob&bench-def} introduces the knapsack problem and how the dynamism is applied to our benchmarks. The baseline evolutionary algorithms are introduced in Section~\ref{sec:baselineEAs}. We provide a theoretical runtime analysis of the baseline approaches in Section~\ref{sec:theo} and an in depth experimental investigation of them in Section~\ref{subec:baseline-results}.
In Section~\ref{sec:NSGAII-SPEA2}, we present NSGA-II and SPEA2, the necessary modifications to apply them on the dynamic knapsack problem, and their experimental results in detail. Finally, a conclusion follows in Section~\ref{sec:conc}.

\section{The Knapsack Problem with Dynamic Constraints}\label{sec:prob&bench-def}
In this section, we define the Knapsack Problem (KP) and further notation used in the rest of this paper. We present how a dynamic change impacts the constraint bound in KP, and introduce the details of benchmarks and the experimental settings used in Sections~\ref{subec:baseline-results} and~\ref{sec:NSGAII-SPEA2}.

\subsection{Problem Definition}
We investigate the performance of different evolutionary algorithms on the KP under dynamic constraint. There are $n$ items with profits $\{p_1,\ldots, p_n\}$ and weights $\{w_1,\ldots, w_n\}$. A solution $x$ is a bit string of $\{0,1\}^n$ which has the overall weight $w(x) = \sum_{i=1}^{n}{w_ix_i}$ and profit $p(x) = \sum_{i=1}^{n}{p_ix_i}$. 
The goal is to find a solution $x^* = \arg \max \{p(x) \mid x \in \{0,1\}^n \wedge w(x) \leq C\}$ of maximum profit whose weight does not exceed the capacity constraint $C$.

We consider two types of this problem based on the consideration of the weights. Firstly, we assume that all the weights are one and uniform dynamic constraint is applied. In this case, the limitation is on the number of items chosen for each solution and the optimal solution is to pick $C$ items with the highest profits. Next, we consider the general case where the profits and weights are integers under linear constraint on the weight.

\subsection{Dynamically Changing Constraints} 
In the following section, the dynamic version of KP used for the experiments is described, and we explain how the dynamic changes occur during the optimization process.

\begin{figure}[ht]
	\centering
	\begin{subfigure}[b]{0.490\textwidth}
		\centering
		\includegraphics[width=\textwidth]{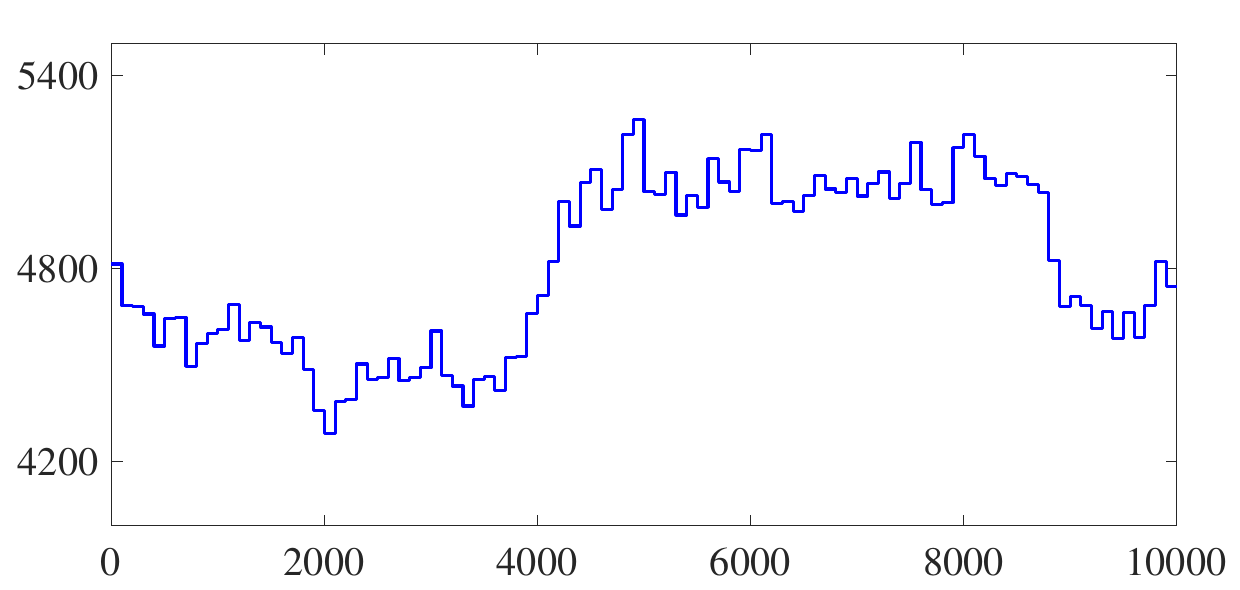}
		\caption[]%
		{\centering{\small $\mathcal{N}(0,100^2)$.}}    
		\label{fig:norm100}
	\end{subfigure}
	\hfill
	\begin{subfigure}[b]{0.490\textwidth}  
		\centering 
		\includegraphics[width=\textwidth]{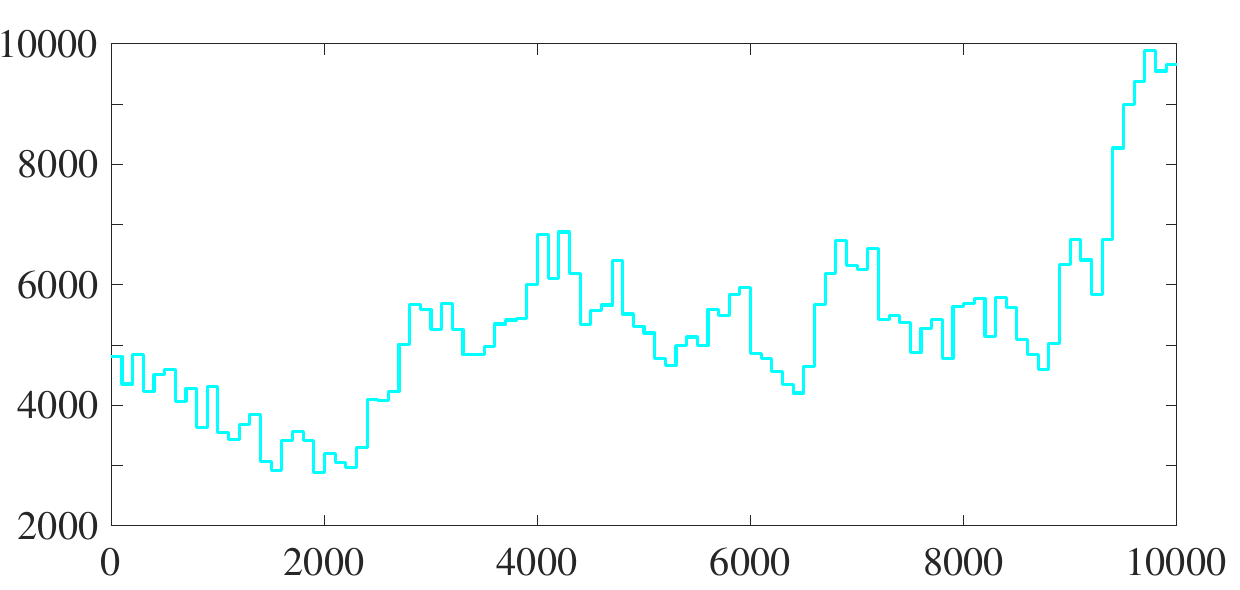}
		\caption[]%
		{\centering{\small $\mathcal{N}(0,500^2)$.}}    
		\label{fig:norm500}
	\end{subfigure}
	\vskip\baselineskip
	\begin{subfigure}[b]{0.48\textwidth}   
		\centering 
		\includegraphics[width=\textwidth]{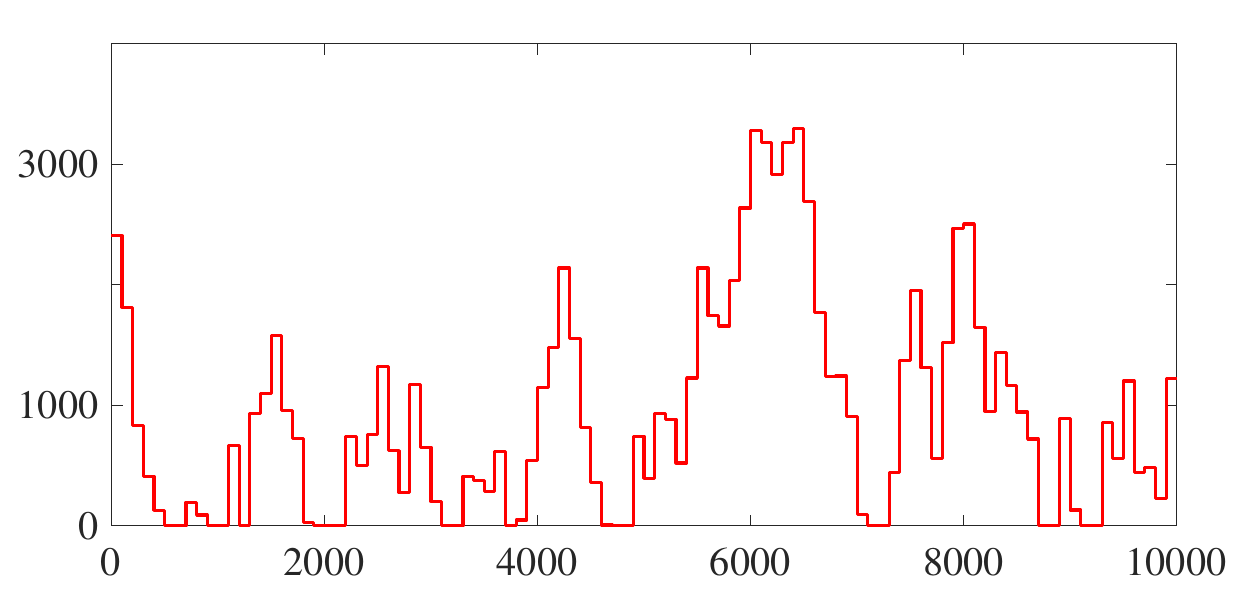}
		\caption[]%
		{\centering{\small $\mathcal{U}(-2000,2000)$.}}    
		\label{fig:uni2000}
	\end{subfigure}
	\quad
	\begin{subfigure}[b]{0.48\textwidth}   
		\centering 
		\includegraphics[width=\textwidth]{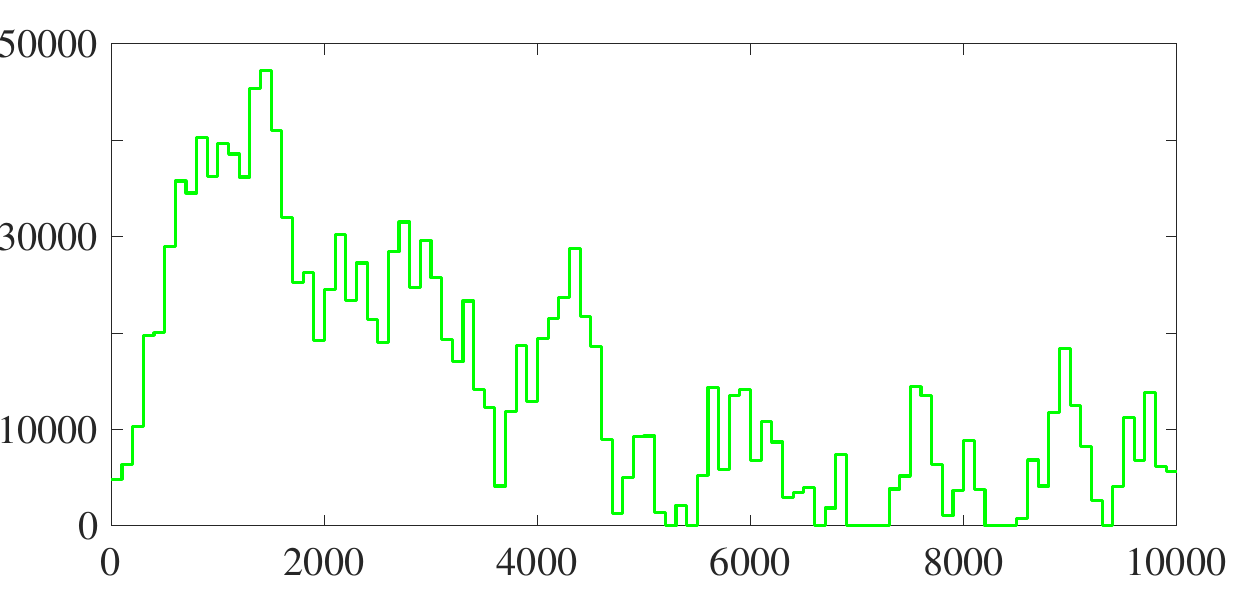}
		\caption[]%
		{\centering{\small $\mathcal{U}(-10000,10000)$.}}    
		\label{fig:uni10000}
	\end{subfigure}
	\caption{Examples for constraint bound $C$ over 10000 generations with $\tau =100$ using uniform and normal distributions. Initial value $C = 4815$. The X axis indicates the value of $C$ and $Y$ axis is the generation number.} 
	\label{fig:distributions}
\end{figure}

In the dynamic version of KP considered in this paper, the capacity dynamically changes during the optimization with a preset frequency factor denoted by $\tau$. A change happens every $\tau$ generations, i.e., the algorithm has $\tau$ generations to find the optimum of the current capacity and to prepare for the next change. In the case of uniformly random changes, the capacity of next interval is achieved by adding a uniformly random value in $[-r, r]$ to $C$. Moreover, we consider another case in which the amount of the changes is chosen from the Gaussian distribution $\mathcal{N}(0, \sigma^2)$. During the process, the capacity is capped at zero and sum of items' weight. Figure~\ref{fig:distributions} illustrates how dynamic changes from different distributions affect the capacity. 
Note that the scales of the subfigures are not the same. For example, the difference between the minimum and the maximum capacity after 100 dynamic changes under $\mathcal{N}(0, 100^2)$ is almost 1000 (Figure~\ref{fig:norm100}) while the capacity reached almost 45000, starting from $4815$, taking dynamic changes under $\mathcal{U}(-10000, 10000)$ (Figure \ref{fig:uni10000}). This indicates that there are different types of challenges, resulting from the dynamic changes that the algorithms must consider.

The combination of different distributions and frequencies brings interesting challenges for the algorithms. In an environment where the constraint changes with a high frequency, the algorithms have less time to find the optimal solution, hence, it is likely that an algorithm which tries to improve only one solution will perform better than another algorithm that needs to optimize among several solutions. On the other hand, the algorithms that pay a lot of attention to the configuration of solutions in objective space, might lose the track of new optimal solution. This is caused by their preference in solutions with better distribution factor, instead searching for feasible solutions that are closer to the capacity constraint. Furthermore, the uniform distribution guarantees upper and lower bounds on the magnitude of the changes. This property could be beneficial for the algorithms which keep a certain number of solutions in each generation, so that they do get ready and react faster after a dynamic change. If the changes happen under a normal distribution, however, there is no strict bound on the value of any particular change, which means it is not easy to predict which algorithms will perform better in this type of environment.
\subsection{Benchmark and Experimental Setting} \label{subsec: bech and exp}
In this section, we present the dynamic benchmarks and the experimental settings. We use eil101 benchmarks, which were originally generated for Traveling Thief Problem \cite{polyakovskiy2014comprehensive} with 100 cities and one item on each city, ignoring the cities and only using the items.
The weights and profits are generated in three different classes. In Uncorrelated (uncorr) instances, the weights and profits are integers chosen uniformly at random within $[1,1000]$. The Uncorrelated Similar Weights (unc-s-w) instances have uniformly distributed random integers as the weights and profits within $[1000, 1010]$ and $[1, 1000]$, respectively. Finally, there is the Bounded Strongly Correlated (bou-s-c) variations which result in the hardest instances and comes from the bounded knapsack problem. The weights are chosen uniformly at random within the interval $[1, 1000]$, and the profits are set according to the weights plus 100. In addition, we set all the weights to one and consider the original profits from the benchmarks. The initial capacity in this version is calculated by dividing the original capacity by the average of the profits.
Dynamic changes add a value to $C$ each $\tau$ generations. Four different situations in terms of frequencies are considered: high frequency changes with $\tau=100$, medium frequency changes with $\tau=1000$, $\tau=5000$ and low frequency changes with $\tau=15000$.

For case where all the weights are set to 1, we only consider the dynamic changes that are chosen uniformly at random within the interval $[-5, 5]$. In the case of the original weights, we investigate two settings. Firstly, the magnitude of changes are generated uniformly at random within the interval $[-r,r]$, where $r=2000, 10000$. In the second setting, the changes are generated by using normal distribution with $\sigma=100$, and $\sigma=500$.

We use the offline errors to compute the performance of the algorithms. In each generation, we record error $e_i=p(x^*_i)-p(x_i)$ where $x^*_i$ and $x_i$ are the optimal solution and the best achieved feasible solution in generation $i$, respectively. If no feasible solution is found in generation $i$, we consider solution $y_i$ that denotes the solution with the smallest constraint violation. Then, the offline error is calculated as $e_i=p(x^*_i)+\nu(y_i)$. Hence, the total offline error for $m$ generations would be $\sum_{i=1}^{m}{e_i}/m$.

The benchmarks for dynamic changes are thirty different instances. Each instance consists of $100000$ changes, as numbers in $[-r, r]$ generated uniformly at random. Similarly, there are thirty other instances with 100000 numbers generated under the normal distribution $\mathcal{N}(0, \sigma^2)$. The algorithms start from the beginning of each file and pick the number of change values from the files. Hence, for each setting, we run the algorithms thirty times with different dynamic change values and record the total offline error of each run.

In order to establish a statistical comparison of the results among different algorithms, we use a multiple comparisons test. In particularity, we focus on the method that compares a set of algorithms. 
For statistical validation we use the Kruskal-Wallis test with $95$\% confidence. Afterwards, we apply the Bonferroni post-hoc statistical procedures that are used for multiple comparisons of a control algorithm against two or more other algorithms.
For more detailed descriptions of the statistical tests we refer the reader to~\cite{Corder09}.

Our results are summarized in the Tables~\ref{tbl:uni_w1}-~\ref{tbl:par_norm}. 
The columns represent the algorithms with the corresponding mean value and standard deviation.
Note, $X^{(+)}$ is equivalent to the statement that the algorithm in the column outperformed algorithm $X$, and $X^{(-)}$ is equivalent to the statement that $X$ outperformed the algorithm in the given column. If the algorithm $X$ does not appear, this means that no significant difference was observed between the algorithms.

\section{Baseline Evolutionary Algorithms}\label{sec:baselineEAs}
In this section, we introduce the baseline evolutionary algorithms considered in this study and present detailed comparison of their performance using the total offline values.

\subsection{Algorithms}

We investigate the performance of three algorithms in this section. The initial solution for all these algorithms is a solution with items chosen uniformly at random. After a dynamic change to constraint $C$ happens, all the algorithms polish their solution sets according to the new capacity and resume the optimization. This process is to consider previous solutions that become infeasible after the change and to remove solutions that are not worth to be kept anymore.

The (1+1)~EA (Algorithm \ref{alg:1+1ea}) flips each bit of the current solution with the probability of $\frac{1}{n}$ as the mutation step. Afterwards, the algorithm chooses between the original solution and the mutated one using the value of the fitness function. Let $p_{\max} = \max_{1\leq i\leq n} p_i$ be the maximum profit among all the items. The fitness function that we use in the (1+1)~EA is as follows:

\begin{algorithm}[t]
	$x \leftarrow$ previous best solution\;
	\While {stopping criterion not met}{
		$y \leftarrow$ flip each bit of $x$ independently with probability of $\frac{1}{n}$\;
		\If{$f_{1+1}(y) \geq f_{1+1}(x)$}{$x \leftarrow y$\;}
	}
	\caption{(1+1)~EA}\label{alg:1+1ea}
\end{algorithm}
$$ f_{1+1}(x) = p(x)-(n\cdot p_{\max}+1) \cdot \nu(x)\text{,}$$
where $\nu(x) = \max \left\{ 0,w(x)-C \right\}$ is the constraint violation of $x$. If $x$ is a feasible solution, then $w(x)\leq C$ and $\nu(x)=0$. Otherwise, $\nu(x)$ is the weight distance of $w(x)$ from $C$.

The algorithm aims to maximize $f_{1+1}$ which consists of two terms. The first term is the total profit of the chosen items and the second term is the penalty applied to infeasible solutions. The amount of penalty guarantees that a feasible solution always dominates an infeasible solution. Moreover, for two infeasible solutions, the one with weight closer to $C$ dominates the other one.

The other algorithm we consider in this paper is a multi-objective evolutionary algorithm (Algorithm~\ref{alg:moea}), which is motivated by a theoretical study on the performance of evolutionary algorithms for the reoptimization of linear functions under dynamic uniform constraints~\cite{DBLP:journals/algorithmica/ShiSFKN19}. In the case of a uniform constraint the weight $w(x)$ of a solution $x$ is given by the number of chosen elements, i.e.,  $w(x)=|x|_1$ holds.
Each fitness of solution $x$ in the bi-objective objective space is a two-dimensional point $f_{MOEA}(x) = (w(x), p(x))$. 
We say solution $y$ dominates solution $x$ w.r.t. $f_{MOEA}$, denoted by $y \succcurlyeq_{MOEA} x$, if $w(y) = w(x) \land f_{(1+1)}(y) \geq f_{(1+1)}(x)$.

\begin{algorithm}[t]
	Update $C$\;
	$S^+ \leftarrow \{z \in S^+ \cup S^- \mid C < w(z)\leq C+\delta\}$\;
	$S^- \leftarrow \{z \in S^+ \cup S^- \mid C-\delta \leq w(z)\leq C\}$\;
	\If {$S^+ \cup S^- = \emptyset$ }{$q\leftarrow$ best found solution before the dynamic change\; 
	}
	\uIf {$C< w(q)\leq C+\delta$}{
		$S^+\leftarrow\{q\}\cup S^+$\;}
	\ElseIf {$C-\delta \leq w(q)\leq C$}{$S^-\leftarrow\{q\}\cup S^-$\;}
	
	\While {a change happens}{
		\eIf{$S^+\cup S^- = \emptyset$}{
			Initialize $S^+$ and $S^-$ by Repair($q$,$\delta$,$C$)\;
		}
		{
			choose $x\in S^+ \cup S^-$ uniformly at random\;
			$y \leftarrow$ flip each bit of $x$ independently with probability $\frac{1}{n}$\; 
			\If{ $(C < w(y) \leq C + \delta) \land (\nexists p\in S^+ : p\succcurlyeq_{MOEA} y)$ }
			{ $S^+\leftarrow (S^+ \cup \{y\})\setminus \{z\in S^+ \mid y\succ_{MOEA} z \} $\;}
			\If{ $(C-\delta \leq w(y) \leq C ) \land (\nexists p\in S^- : p\succcurlyeq_{MOEA} y)$ }
			{ $S^-\leftarrow (S^- \cup \{y\})\setminus \{z\in S^- \mid y\succ_{MOEA} z \} $\;}
		}
	}
	\caption{\EA}\label{alg:moea}
\end{algorithm}

\begin{algorithm}[t]
	
	\SetKwInOut{Input}{input}\SetKwInOut{Output}{output}
	\Input{Initial solution $q$, $\delta$, $C$}
	\Output{ $S^+$ and $S^-$ such that $|S^+ \cup S^-|=1$}
	\While{$S^+\cup S^- = \emptyset$}{
		$y \leftarrow$ flip each bit of $q$ independently with probability of $\frac{1}{n}$\;
		\If{$f_{1+1}(y) \geq f_{1+1}(q)$}{$q \leftarrow y$\;
			\uIf {$C< w(q)\leq C+\delta$}{
				$S^+\leftarrow\{q\}\cup S^+$\;}
			\ElseIf {$C-\delta \leq w(q)\leq C$}{$S^-\leftarrow\{q\}\cup S^-$\;}
		}
	}
	\caption{Repair}\label{alg:repair}
\end{algorithm}
According to the definition of $\succcurlyeq_{MOEA}$, two solutions are comparable only if they have the same weight and reflects the same approach as investigated in \cite{DBLP:journals/algorithmica/ShiSFKN19} for the case where each element contributes a weight of $1$. Note that if $x$ and $y$ are infeasible and comparable, then the one with higher profit dominates. \EA uses a parameter denoted by $\delta$, which determines the maximum number of individuals that the algorithm is allowed to store around the current $C$. For any weight in $[C-\delta, C+\delta]$, \EA keeps one solution. The algorithm prepares for the dynamic changes by storing nearby solutions, even if they are infeasible as they may become feasible after the next change. A large $\delta$, however, causes a large number of solutions to be kept, which reduces the probability of choosing a particular one. Since the algorithm chooses only one solution to mutate in each iteration, this affects \EA's performance in finding the optimal solution. 

After each dynamic change, \EA updates the sets of solutions. If a change occurs such that all the current stored solutions are outside of the storing range, namely $[C-\delta, C+\delta]$, then the algorithm considers the previous best solution as the initial solution and uses the Repair function (Algorithm~\ref{alg:repair}), which behaves similar to (1+1)~EA, until a solution with weight distance at most $\delta$ from $C$ is found. Note that if all the solutions in the previous interval were infeasible, the best previous solution is the one with the lowest weight.

To address the slow rate of improvement of \EA caused by a large $\delta$, we use the standard definition of dominance in multi-objective optimization-- i.e., solution $y$ dominates solution $x$, denoted by $y\succcurlyeq_{\text{\textit{MOEA}}_D}x$, if $w(y)\leq w(x) \land p(y)\geq p(x)$. This new algorithm, called \EAd, is obtained by replacing $\succcurlyeq_{MOEA}$ with $\succcurlyeq_{\text{\textit{MOEA}}_D}$ in lines 16-19 of Algorithm~\ref{alg:moea}. It should be noticed that if $y$ is an infeasible solution then it is only compared with other infeasible solutions and if $y$ is feasible it is only compared with other feasible solutions. 
\EAd keeps fewer solutions than \EA and the overall quality of the kept solutions is higher, since they are not dominated by any other solution in the population.



\section{Theoretical Analysis of Baseline Approaches}
\label{sec:theo}
The approaches introduced in the previous section are motivated by recent theoretical investigations in the area of runtime analysis carried out in \cite{DBLP:journals/algorithmica/ShiSFKN19,DBLP:journals/algorithmica/ShiSFKN20}.
In the following, we provide some additional theoretical insights into the use of approaches studied mainly experimentally in this paper.
It has already been shown theoretically in \cite{DBLP:journals/algorithmica/ShiSFKN19,DBLP:journals/algorithmica/ShiSFKN20} that the use of a multi-objective formulation yields to obtain better runtime bounds for a variant of \EA than for the (1+1)~EA. The runtime bounds shown there for the case of a uniform constraint only show a small difference as even the (1+1)~EA has a polynomial expected optimization time.
In following, we provide simple examples based on trap functions that show an exponential optimization time for the (1+1)EA while \EA and \EAd with the right choice of $\delta$ have a small polynomial expected optimization time. Furthermore, we show that a too small choice of $\delta$ leads to an exponential optimization time for \EA and \EAd.

There are two significant benefits of working with a multi-objective formulation which stores a range of trade-offs with respect to the weight and profits of a solution when working with dynamically changing constraints. 
The first one is that if the population contained already good solutions for a range of weights around the constraint bound then it is likely to have already a good solution for the bound after the dynamic change if the change in constraint bound value has been relatively small. The second benefit lies in the ability of the multi-objective formulation as well as it has been observed in many Pareto optimization approaches. Even if a solution has been good before a dynamic change has occurred, then a small change in the value of the constraint bound might require a structurally completely different solution. As we will show in the following, even simple instances in the dynamic setting can create local optima which are hard to escape after a dynamic change occurred.

We consider the trap instance $T$ consisting of $n$ items where $p_i=w_i=1$, $1 \leq i \leq n-1$ and $p_n=w_n=n$. All search points are Pareto optimal as the weight equals its profit for each item.
In order to show a lower bound for the (1+1)~EA, we assume that the algorithm has obtained the optimal solution for capacity $C=n-1$ and the capacity is changed afterwards to $C^*=n$.

\begin{theorem}
	The time until the (1+1)~EA has produced from the optimal solution $x^*$ for capacity $C=n-1$ for instance $T$, an optimal solution for capacity $C^*=C+1=n$ after the dynamic change has occurred is at least $n^{n/2}$ with probability at least $1-n^{-n/2}$.
\end{theorem}

\begin{proof}
	The optimal solution $x^*$ for capacity $C=n-1$ consists of the first $n-1$ items whereas the optimal solution $y^*$ for capacity $C^*=n$ consists of the last item only. Having obtained the solution $x^*$ the (1+1)~EA only accepts $y^*$ as this is the only solution $y$ with $p(y) \geq p(x^*)$ and $w(y) \leq C$. The probability to produce from $x^*$ the solution $y^*$ is $n^{-n}$ as all bits of $x^*$ have to flip at the same time. Hence, the expected optimization time is $n^n$, and the optimization time is at least $n^{n/2}$ with probability at least $1-n^{-n/2}$ using the union bound.
\end{proof}

In the following theorem we show that \EA and \EAd compute an optimal solution for instance $T$ for every possible capacity $C \in \{0, \ldots, 2n-1\}$ independently of the starting population. This includes the population that contains only the optimal solution for capacity $n-1$ which leads to the exponential lower bound for (1+1)~EA.

Note for any solution $x$ of instance $T$, $p(x)=w(x)$ holds. This implies that two solutions $x$ and $y$ with $w(x) \not = w(y)$ do not dominate each other and we have $y \succcurlyeq_{MOEA} x$ iff $y \succcurlyeq_{MOEA_D} x$. Therefore, $\EA$ and $\EAd$ perform identically on instance $T$.

\begin{theorem}
	Starting with an arbitrary population, the expected time until \EA and \EAd  have produced an optimal solution for any capacity $C^* \in \{0, \ldots, 2n-1\}$ for instance $T$ is $O(n^2 \log n)$ if $\delta \geq n$.
\end{theorem}

\begin{proof}
	All search points take on values $w(x)=p(x) \in \{0, \ldots, 2n-1\}$ which implies that the number of trade-off solutions is at most $2n$ and the population size is always upper bounded by $2n$.
	
	We consider two cases. First assume that $0 \leq C \leq n-1$. Then the optimal solution selects exactly $C$ items from the set of the first $n-1$ items.
	As $\delta \geq n$, search points with different values $p(x) \in \{0, \ldots, n-1\}$ are accepted. In particular, each feasible solution $x$ has a profit  $p(x) \in \{C-\delta, \ldots, C+ \delta\}$. If \EA has not obtained yet a solution $x$ with $p(x) \in \{C- \delta, C+ \delta\}$, then the population size is $1$, the current solution $x$ is infeasible and the (1+1)~EA is evoked to produce a feasible solution. The expected time to reach a solution with $w(x) \leq n-1$ is $O(n)$ as removing item $n$ from the current search point (if present) leads to such a solution and happens with probability $\Omega(1/n)$. If the current solution has weight $w(x) \leq n-1$, then there are $w(x)=|x|_1$ $1$-bits in $x$ that can be flipped to reduce the weight. Using fitness level arguments~(see for example Section 4.2.1 in \cite{DBLP:books/daglib/0025643}) with respect to the values $w(x)$, $1 \leq w(x) \leq n-1$, and taking into account that the population size is $O(n)$ the search point $0^n$ is produced within $O(n^2 \log n)$ steps. This solution is feasible as $C \geq 0$. In order to obtain an optimal solution $x^*$ with $p(x^*)=C$, we consider the feasible solution $x$ with the largest profit in the population. This solution is selected for mutation in the next iteration with probability $\Omega(1/n)$ and any of the $(n-1)-|x|_1$ bits at the first $n-1$ positions currently set to $0$ can flip to increase the value of $p(x)$.
	Such an improvement happens with probability $\Omega(((n-1)-|x|_1)/(en^2))$ in the next step.
	Using fitness level argument with respect to the different profit values $p(x)$, $0 \leq p(x) \leq C$, the expected time to obtain a solution $x^*$ with $p(x^*)=C$ is $O(n^2 \log n)$.
	
	
	We now assume that $n \leq C \leq 2n-1$. Then the optimal solution has to select item $n$ and $C-n$ items from the first $n-1$ items. We use similar arguments as in the previous case to show the $O(n^2 \log n)$ upper bound.
	We first consider the time to obtain a solution $x$ with weight $w(x) \in \{C - \delta, \ldots, C+\delta\}$. Any solution containing the item $n$ has a weight in that range. If a solution within the range has not been obtained yet, then a solution within that range is achieved in expected time $O(n)$ as only the bit corresponding to item $n$ needs to be flipped with happens with probability $\Omega(1/n)$. Once item $n$ is contain in the population, a solution with profit $n$ is obtained in expected time $O(n^2 \log n)$ which is obtained by considering always the individual $x$ in the population with the smallest profit $p(x) >n$, selecting this for mutation and flipping any of the $p(x)-n$ $1$-bits corresponding to the first $n-1$ items. The solution consisting of item $n$ is feasible as $C \geq n$. In order to obtain a solution with profit $C$, we use again fitness level arguments with respect to the number of elements from the set of the first $n-1$ elements (in addition to element $n$). The probability to increase the profit of the feasible solution $x$ with the high profit in the population is at least $(n- |x|_1)/(en^2)$. Using fitness level arguments with respect to the different profit values an optimal solution for capacity $C$ is obtained in expected time $O(n^2 \log n)$.
\end{proof}

The previous theorem shows that the use of the population with a suitable large value of $\delta$ in \EA and \EAd can lead to an exponential speed up compared to the (1+1)~EA.

We now show a situation where \EA and \EAd encounter an exponential re-optimization time if the change of the constraint bound is only by one larger than $\delta$. It should be noted that we already observed this phenomenon when comparing (1+1)~EA to \EA as (1+1)~EA is equivalent to \EA with $\delta=0$.

Consider again the instance $T$. We set $C=0.75n$, $\delta=0.25n-1$ and assume that the population contains for each value in $\{C-\delta, C+\delta\}=\{0.5n+1, \ldots, n-1\}$ an optimal solution.

\begin{theorem}
	Starting with an optimal population for instance $T$, $C=0.75n$ and $\delta=0.25n-1$, the time until \EA and \EAd  have produced a new accepted solution after changing the capacity to $C^*=n$ is at least $e^{n/4}$ with probability $1-e^{-n/4}$.
\end{theorem}
\begin{proof}
	Changing the bound to $C^*=n$ implies that the population is updated such that it consists of optimal solutions for the values $\{0.75n+1, \ldots, n-1\}$. 
	After the change only solutions with weights in $\{0.75n+1, \ldots, 1.25n-1\}$ are accepted. In order to produce from any solution with weight in $\{0.75n+1, \ldots, n-1\}$ a solution with weight in $\{n, \ldots, 1.25n-1\}$, element $n$ has to be introduced and at least 
	$$
	(0.75n+1) +n - (1.25n-1)=0.75n+1 - (0.25n-1) > 0.5n$$ elements from the first $n-1$ elements need to be removed at the same time. The probability for this to happen in a single mutation step is at most $e^{-n/2}$ as at least $n/2$ bits need to flip at the same time. The time for such a step to occur is therefore at least $e^{n/4}$ with probability $1-e^{-n/4}$ using the union bound.
\end{proof}

\section{Experimental Results for Baseline Approaches}\label{subec:baseline-results}
We now investigate the introduced algorithms from an experimental perspective.
In this section we describe the initial settings of the algorithms and analyze their performance using statistical tests.
The initial solution for all the algorithms is a pack of items which are chosen uniformly at random. 
Each algorithm initially runs for 10000 generations without any dynamic change. After this, the first change is introduced, and the algorithms run one million further generations with dynamic changes in every $\tau$ generations.
For \EA and \EAd, it is necessary to initially provide a value for $\delta$. These algorithms keep at most $\delta$ feasible solutions and $\delta$ infeasible solutions, to help them efficiently deal with a dynamic change. When the dynamic changes come from $\mathcal{U}(-r, r)$, it is known that the capacity will change by most $r$. Hence, we set $\delta=r$. In case of changes from $\mathcal{N}(0, \sigma^2)$, $\delta$ is set to $2\sigma$, since $95\%$ of values will be within $2\sigma$ of the mean value. Note that a larger $\delta$ value increases the population size of the algorithms and there is a trade-off between the size of the population and the speed of algorithm in reacting to the next change.

\subsection{Dynamic Uniform Constraint}\label{sec:w-1}
In this section, we validate the theoretical results against the performance of (1+1)~EA and Multi-Objective Evolutionary Algorithm.~\cite{DBLP:journals/algorithmica/ShiSFKN19} state that the multi-objective approach performs better than (1+1)~EA in re-optimizing the optimal solution of dynamic KP under uniform constraint. Although the \EA that we used in this experiment is not identical to the multi-objective algorithm studied previously by~\cite{DBLP:journals/algorithmica/ShiSFKN19} and they only considered the re-optimization time, the experiments show that multi-objective approaches outperform (1+1)~EA in the case of uniform constraints (Table \ref{tbl:uni_w1}). 
\begin{table}[t]
\centering
\scriptsize
\setlength{\tabcolsep}{5pt}

\caption{The mean, standard deviation values and statistical tests of the offline error for (1+1)~EA, \EA, \EAd based on the uniform distribution with all the weights as one.}
\label{tbl:uni_w1}
\begin{tabular}{llllrrlrrlrrl}
        & \multicolumn{1}{c}{$n$} & \multicolumn{1}{c}{$r$} & \multicolumn{1}{c}{$\tau$} & \multicolumn{3}{c}{(1+1)~EA (1)}                                                          & \multicolumn{3}{c}{\EA (2)}                                                 & \multicolumn{3}{c}{\EAd (3)}                                               \\
        &                       &                       &                            & \multicolumn{1}{c}{mean} & \multicolumn{1}{c}{st} & \multicolumn{1}{c}{stat}              & \multicolumn{1}{c}{mean} & \multicolumn{1}{c}{st} & \multicolumn{1}{c}{stat} & \multicolumn{1}{c}{mean} & \multicolumn{1}{c}{st} & \multicolumn{1}{c}{stat} \\ \hline
uncor   & 100                   & \multicolumn{1}{c}{5} & 100                        & 4889.39                & 144.42               & $2^{(-)}$,$3^{(-)}$                    & 1530.00                & 120.76               & $1^{(+)}$                & \textbf{1486.85}                & 123.00               & $1^{(+)}$                \\
        & 100                   & \multicolumn{1}{c}{5} & 1000                       & 1194.23                & 86.52                & $2^{(-)}$,$3^{(-)}$                   & \textbf{44.75 }                 & 8.96                 & $1^{(+)}$               & 46.69                  & 8.51                 & $1^{(+)}$                \\ \hline
unc-s-w & 100                   & \multicolumn{1}{c}{5} & 100                        & 4990.80                & 144.87               & $2^{(-)}$,$3^{(-)}$ 
& 1545.36                & 115.15               & $1^{(+)}$                & \textbf{1500.07}                & 106.70               & $1^{(+)}$                \\
        & 100                   & \multicolumn{1}{c}{5} & 1000                       & 1160.23                & 130.32               & $2^{(-)}$,$3^{(-)}$                   & \textbf{41.90}                  & 6.13                 & $1^{(+)}$                & 43.06                  & 7.22                 & $1^{(+)}$                \\ \hline
bou-s-c & 100                   & \multicolumn{1}{c}{5} & 100                        & 13021.98               & 780.76               & $2^{(-)}$,$3^{(-)}$                    & 4258.53                & 580.77               & $1^{(+)}$               & \textbf{4190.55}                & 573.13               & $1^{(+)}$               \\
        & 100                   & \multicolumn{1}{c}{5} & 1000                       & 3874.76                & 911.50               & $2^{(-)}$,$3^{(-)}$                   & 177.62                 & 83.16                & $1^{(+)}$                & \textbf{175.14 }                & 80.73                & $1^{(+)}$               \\ \hline

\end{tabular}
\end{table}
An important reason for this remarkable performance is the relation between optimal solutions in different weights. In this type of constraint, the difference between the optimal solution of weight $w$ and $w+1$ is one item. As a result of this, keeping non-dominated solutions near the constrained bound helps the algorithm to find the current optimum more efficiently and react faster after a dynamic change. 

Furthermore, according to the results, there is no significant difference between using \EA and \EAd in this type of KP. Considering the experiments in Section~\ref{subsec:dyn_linear}, a  possible reason is that the size of population in \EA remains small when weights are one. Hence, \EAd, which stores fewer items because of its dominance definition, has no advantage in this manner anymore. In addition, the constraint is actually on the number of the items. 

\subsection{Dynamic Linear Constraint}\label{subsec:dyn_linear}
In this section, we consider the same algorithms in more difficult environments where weights are arbitrary under dynamic linear constraint. As it is shown in Section~\ref{sec:w-1}, the multi-objective approaches outperform (1+1)~EA in the case that weights are one. Now we try to answer the question: Does the relationship between the algorithms hold when the weights are arbitrary?

The data in Table~\ref{tbl:uni_wg} shows the experimental results in the case of dynamic linear constraints and changes under a uniform distribution. It can be observed that (as expected) the mean of errors decreases as $\tau$ increases. Larger $\tau$ values give more time to the algorithm to get closer to the optimal solution. Moreover, starting from a solution which is near to the optimal for the previous capacity, can help to speed up the process of finding the new optimal solution in many cases.

We first consider the results of dynamic changes under the uniform distribution. We observe in Table~\ref{tbl:uni_wg} that unlike with uniform constraint, in almost all the settings, \EA has the worst performance of all the algorithms. The first reason for this might be that, in case of the uniform constraints, the magnitude of a capacity change is equal to the Hamming distance of optimal solutions before and after the change. In other words, when weights are one, we can achieve the optimal solution for weight $w$ by adding an item to the optimal solution for weight $w-1$ or by deleting an item from the optimal solution for $w+1$. However, in case of arbitrary weights, the optimal solutions of weight $w$ and $w+d$ could have completely different items, even if $d$ is small. Another reason could be the effect of having a large population. A large population may cause the optimization process to take longer and it could get worse because of the definition of $\succcurlyeq_{MOEA}$, which only compares solutions with equal weights. If $s$ is a new solution and there is no solution with $w(s)$ in the set of existing solutions, \EA keeps $s$ whether $s$ is a good solution or not, i.e., regardless of whether it is really a non-dominated solution or whether there exist another solution with lower weight and higher profit in the set. This comparison also does not consider whether $s$ has any good properties to be inherited by the next generation. For example, \EA generates $s$ that includes items with higher weights and lower profits. Since it might have a unique weight, \EA keeps it in the population. Putting $s$ in the set of solutions decreases the probability of choosing all other solutions, even those solutions that are very close to the optimal solution. As it can be seen in the Table~\ref{tbl:uni_wg}, however, there is only one case in which \EA beats (1+1)~EA: when the weights are similar, and the magnitude of changes are small (2000), which means the population size is also small (in comparison to 10000), and finally $\tau$ is at its maximum to let the \EA to use its population to optimize the problem. 

Although \EA does not perform very well in instances with general weights, the multi-objective approach with a better defined dominance, \EAd, does outperform (1+1)~EA in many cases. We compare the performance of (1+1)~EA and \EAd below.

\begin{table}[t]
\centering
\scriptsize
\setlength{\tabcolsep}{2.8pt}

\caption{The mean, standard deviation values and statistical tests of the offline error for (1+1)~EA, \EA, \EAd based on the uniform distribution.}
\label{tbl:uni_wg}
\begin{tabular}{llllrrlrrlrrl}
        & \multicolumn{1}{c}{$n$} & \multicolumn{1}{c}{$r$} & \multicolumn{1}{c}{$\tau$} & \multicolumn{3}{c}{(1+1)~EA (1)}                                        & \multicolumn{3}{c}{\EA (2)}                                                 & \multicolumn{3}{c}{\EAd (3)}                                               \\
        &                            &                       &                            & \multicolumn{1}{c}{mean} & \multicolumn{1}{c}{st} & \multicolumn{1}{c}{stat} & \multicolumn{1}{c}{mean} & \multicolumn{1}{c}{st} & \multicolumn{1}{c}{stat} & \multicolumn{1}{c}{mean} & \multicolumn{1}{c}{st} & \multicolumn{1}{c}{stat} \\ \hline
uncor   & 100                        & 2000                  & 100                        & 5564.37               & 463.39               & $2^{(+)}$,$3^{(-)}$      & 11386.40                & 769.77               & $1^{(-)}$,$3^{(-)}$      & \textbf{3684.26}                & 525.50               & $1^{(+)}$,$2^{(+)}$      \\
        & 100                        & 2000                  & 1000                       & 2365.56                & 403.64               & $2^{(+)}$,$3^{(-)}$      & 7219.17               & 587.50              & $1^{(-)}$,$3^{(-)}$      & \textbf{776.14}                & 334.69               & $1^{(+)}$,$2^{(+)}$      \\
        & 100                        & 2000                  & 5000                       & 1415.42               & 167.08               & $2^{(+)}$,$3^{(-)}$      & 3598.29               & 420.12              & $1^{(-)}$,$3^{(-)}$      & \textbf{270.90}                 & 121.43               & $1^{(+)}$,$2^{(+)}$      \\
        & 100                        & 2000                  & 15000                      & 914.55                 & 102.82              & $2^{(+)}$,$3^{(-)}$       & 2004.16               & 368.82              & $1^{(-)}$,$3^{(-)}$      & \textbf{88.80}                 & 43.98               & $1^{(+)}$,$2^{(+)}$      \\ \hline
unc-s-w & 100                        & 2000                  & 100                        & 3128.43                & 188.36               & $2^{(+)}$,$3^{(-)}$      & 5911.11                & 534.24              & $1^{(-)}$,$3^{(-)}$      & \textbf{2106.45}                & 249.28              & $1^{(+)}$,$2^{(+)}$      \\
        & 100                        & 2000                  & 1000                       & 606.14                 & 99.23               & $2^{(+)}$,$3^{(-)}$      & 1564.23                & 619.97               & $1^{(-)}$,$3^{(-)}$      & \textbf{302.34}                & 24.60                & $1^{(+)}$,$2^{(+)}$      \\
        & 100                        & 2000                  & 5000                       & 147.55                 & 31.80               & $3^{(-)}$                & 174.23                 & 95.98                & $3^{(-)}$                & \textbf{60.94 }                 & 9.12                & $1^{(+)}$,$2^{(+)}$      \\
        & 100                        & 2000                  & 15000                      & 64.65                  & 17.13                & $2^{(-)}$,$3^{(-)}$      & 40.66                  & 15.51               & $1^{(+)}$,$3^{(-)}$      & \textbf{19.26}                 & 4.04                 & $1^{(+)}$,$2^{(+)}$      \\ \hline
bou-s-c & 100                        & 2000                  & 100                        & 3271.07               & 266.54               & $2^{(+)}$                & 5583.53                & 337.81               & $1^{(-)}$,$3^{(-)}$      & \textbf{3036.97}                & 297.33              & $2^{(+)}$                \\
        & 100                        & 2000                  & 1000                       & 1483.01                & 85.14                & $2^{(+)}$,$3^{(-)}$      & 2639.16                & 106.47               & $1^{(-)}$,$3^{(-)}$      & \textbf{617.92}                & 186.35               & $1^{(+)}$,$2^{(+)}$      \\
        & 100                        & 2000                  & 5000                       & 796.77                 & 89.80                & $2^{(+)}$,$3^{(-)}$      & 1256.62                & 118.27               & $1^{(-)}$,$3^{(-)}$      & \textbf{251.41}                 & 109.58               & $1^{(+)}$,$2^{(+)}$      \\
        & 100                        & 2000                  & 15000                      & 538.45                 & 66.98                & $2^{(+)}$,$3^{(-)}$      & 687.95                 & 116.91               & $1^{(-)}$,$3^{(-)}$      & \textbf{104.27}                 & 61.06                & $1^{(+)}$,$2^{(+)}$      \\ \hline
uncor   & 100                        & 10000                 & 100                        & \textbf{10256.72}               & 210.51               & $2^{(+)}$,$3^{(+)}$      & 16278.97               & 248.43               & $1^{(-)}$,$3^{(-)}$      & 11038.07               & 236.91              & $1^{(-)}$,$2^{(+)}$      \\
        & 100                        & 10000                 & 1000                       & 3604.18                & 285.73               & $2^{(+)}$                & 13340.20               & 704.32               & $1^{(-)}$,$3^{(-)}$      & \textbf{3508.51}               & 473.42               & $2^{(+)}$                \\
        & 100                        & 10000                 & 5000                       & 1607.78                & 278.60               & $2^{(+)}$,$3^{(-)}$      & 10614.45              & 1660.32              & $1^{(-)}$,$3^{(-)}$      & \textbf{1183.52}                & 411.83               & $1^{(+)}$,$2^{(+)}$      \\
        & 100                        & 10000                 & 15000                      & 987.64                 & 219.53               & $2^{(+)}$,$3^{(-)}$      & 8006.35                & 1612.20              & $1^{(-)}$,$3^{(-)}$      & \textbf{566.69}                 & 219.54                & $1^{(+)}$,$2^{(+)}$      \\ \hline
unc-s-w & 100                        & 10000                 & 100                        & \textbf{7192.82}                & 153.93               & $2^{(+)}$,$3^{(+)}$      & 12617.69               & 318.23               & $1^{(-)}$,$3^{(-)}$      & 8057.44                & 274.17               & $1^{(-)}$,$2^{(+)}$      \\
        & 100                        & 10000                 & 1000                       & 1846.43                & 115.23               & $2^{(+)}$                & 6981.81                & 768.78               & $1^{(-)}$,$3^{(-)}$      & \textbf{1743.12}                & 364.38               & $2^{(+)}$                \\
        & 100                        & 10000                 & 5000                       & 539.39                 & 65.39                & $2^{(+)}$                & 3488.28               & 819.51               & $1^{(-)}$,$3^{(-)}$      & \textbf{519.63}                 & 175.22               & $2^{(+)}$                \\
        & 100                        & 10000                 & 15000                      & 208.73                 & 36.91                & $2^{(+)}$                & 1525.23                & 306.72               & $1^{(-)}$,$3^{(-)}$      & \textbf{201.97}                 & 79.28                & $2^{(+)}$                \\ \hline
bou-s-c & 100                        & 10000                 & 100                        & \textbf{7187.80}                & 122.59               & $2^{(+)}$,$3^{(+)}$      & 15111.38               & 231.53               & $1^{(-)}$,$3^{(-)}$      & 12736.55               & 229.48               & $1^{(-)}$,$2^{(+)}$      \\
        & 100                        & 10000                 & 1000                       & \textbf{2282.81}                & 219.24               & $2^{(+)}$,$3^{(+)}$      & 8301.43                & 569.90               & $1^{(-)}$,$3^{(-)}$                         & 3575.26                & 550.54               & $1^{(-)}$,$2^{(+)}$      \\
        & 100                        & 10000                 & 5000                       & \textbf{1370.48}                & 250.59               & $2^{(+)}$                & 5248.40                & 1045.78              & $1^{(-)}$,$3^{(-)}$      & 1472.19                & 493.88               & $2^{(+)}$                \\
        & 100                        & 10000                 & 15000                      & \textbf{955.38}                 & 133.33               & $2^{(+)}$                & 3852.07                & 752.84               &  $1^{(-)}$,$3^{(-)}$                        & 977.41                 & 397.75               & $2^{(+)}$               
\end{tabular}
\end{table}
When changes are smaller, it can be seen in Table~\ref{tbl:uni_wg} that the mean of offline errors of \EAd is smaller than of the (1+1)~EA. The dominance of \EAd is such it only keeps the dominant solutions. When a new solution is found, the algorithm compares it to all of the population, removes solutions that are dominated by it and keeps it only if it is not dominated by any other solution in the population. This process improves the quality of the solutions by increasing the probability of keeping a solution beneficial to future generations. Moreover, it reduces the size of the population significantly. Large changes to the capacity, however, makes the \EAd keep more individuals, and then (1+1)~EA may perform better than \EAd.

When $r=10000$, \EAd does not have significantly better results in all cases unlike in the case of $r=2000$, and in most of the situations it performs as well as (1+1)~EA. In all high frequency conditions where $\tau = 100$, the (1+1)~EA has better performance. It may be caused by \EAd needing more time to optimize a population with a larger size. Moreover, when the magnitude of changes is large, it is more likely that a new change will force \EAd to remove all of its stored individuals and start from scratch.

We now study the experimental results that come from considering the dynamic changes under the normal distribution (Table~\ref{tbl:norm_WG}). The results confirm that (1+1)~EA is faster when changes are more frequent. When the variation of changes is small, 
in most of the cases \EAd has been the best algorithm in terms of performance and \EA has been the worst.

\begin{table}[t]
\centering
\scriptsize
\setlength{\tabcolsep}{3.2pt}

\caption{The mean, standard deviation values and statistical tests of the offline error for (1+1)~EA, \EA, \EAd based on the normal distribution.}
\label{tbl:norm_WG}
\begin{tabular}{llllrrlrrlrrl}
        & \multicolumn{1}{c}{$n$} & \multicolumn{1}{c}{$\sigma$} & \multicolumn{1}{c}{$\tau$} & \multicolumn{3}{c}{(1+1)~EA (1)}                                             & \multicolumn{3}{c}{\EA (2)}                                                 & \multicolumn{3}{c}{\EAd (3)}                                               \\
        &                       &                              &                            & \multicolumn{1}{c}{mean} & \multicolumn{1}{c}{st} & \multicolumn{1}{c}{stat} & \multicolumn{1}{c}{mean} & \multicolumn{1}{c}{st} & \multicolumn{1}{c}{stat} & \multicolumn{1}{c}{mean} & \multicolumn{1}{c}{st} & \multicolumn{1}{c}{stat} \\ \hline
uncor   & 100                   & 100                          & 100                        & \textbf{2714.72}                & 106.06               & $2^{(+)}$,$3^{(+)}$      & 9016.83                & 2392.48              & $1^{(-)}$,$3^{(-)}$      & 4271.09               & 789.94               & $1^{(-)}$,$2^{(+)}$      \\
        & 100                   & 100                          & 1000                       & 1386.66                & 97.11                & $2^{(+)}$,$3^{(-)}$      & 3714.89                & 737.11               & $1^{(-)}$,$3^{(-)}$      & \textbf{412.89}                 & 27.25                & $1^{(+)}$,$2^{(+)}$      \\
        & 100                   & 100                          & 5000                       & 801.54                & 73.67                & $2^{(+)}$,$3^{(-)}$      & 1266.35                & 119.25              & $1^{(-)}$,$3^{(-)}$      & \textbf{108.28}                 & 14.22               & $1^{(+)}$,$2^{(+)}$      \\
        & 100                   & 100                          & 15000                      & 549.71                 & 78.98               & $2^{(+)}$,$3^{(-)}$       & 749.86                 & 148.03               & $1^{(-)}$,$3^{(-)}$      & \textbf{61.93}                 & 17.03               & $1^{(+)}$,$2^{(+)}$      \\ \hline
unc-s-w & 100                   & 100                          & 100                        & \textbf{412.24}                 & 111.07               & $2^{(+)}$,$3^{(+)}$      & 1979.65                & 914.35               & $1^{(-)}$                & 1904.09               & 877.55               & $1^{(-)}$                \\
        & 100                   & 100                          & 1000                       & \textbf{85.55}                 & 23.13                & $2^{(+)}$,$3^{(+)}$      & 1566.54                & 409.32              & $1^{(-)}$                & 1482.37                & 391.75              & $1^{(-)}$                \\
        & 100                   & 100                          & 5000                       & \textbf{36.94}                  & 13.61                & $2^{(+)}$,$3^{(+)}$       & 1414.66                & 448.78               & $1^{(-)}$                & 1322.35               & 414.27               & $1^{(-)}$                \\
        & 100                   & 100                          & 15000                      & \textbf{29.14}                  & 19.70               & $2^{(+)}$,$3^{(+)}$      & 1237.67                & 665.27               & $1^{(-)}$                & 1137.80                & 648.73               & $1^{(-)}$                \\ \hline
bou-s-c & 100                   & 100                          & 100                        & \textbf{1491.36}                & 260.72               & $2^{(+)}$,$3^{(+)}$      & 4625.49                & 1302.52              & $1^{(-)}$,$3^{(-)}$      & 2903.77                & 717.92              & $1^{(-)}$,$2^{(+)}$      \\
        & 100                   & 100                          & 1000                       & 736.10                 & 53.99                & $2^{(+)}$,$3^{(-)}$      & 1748.61                & 189.94               & $1^{(-)}$,$3^{(-)}$      & \textbf{312.88}                 & 35.52                & $1^{(+)}$,$2^{(+)}$      \\
        & 100                   & 100                          & 5000                       & 446.94                 & 39.36                & $2^{(+)}$,$3^{(-)}$      & 640.60                 & 91.29                & $1^{(-)}$,$3^{(-)}$      & \textbf{101.21}                & 17.47               & $1^{(+)}$,$2^{(+)}$      \\
        & 100                   & 100                          & 15000                      & 337.85                 & 40.44                & $2^{(+)}$,$3^{(-)}$      & 469.16                 & 93.99                & $1^{(-)}$,$3^{(-)}$      & \textbf{70.16}                  & 22.26                & $1^{(+)}$,$2^{(+)}$      \\ \hline
uncor   & 100                   & 500                          & 100                        &        4013.84         &        699.56         &        $2^{(+)}$,$3^{(-)}$          &          10133.28         &        1128.57         &        $1^{(-)}$,$3^{(-)}$          &          \textbf{2469.58}         &        649.04         &        $1^{(+)}$,$2^{(+)}$       \\
        & 100                   & 500                          & 1000                       &        1991.43         &        163.25         &        $2^{(+)}$,$3^{(-)}$          &          5205.30         &        635.00         &        $1^{(-)}$,$3^{(-)}$          &          \textbf{511.58}         &        187.21         &        $1^{(+)}$,$2^{(+)}$       \\
        & 100                   & 500                          & 5000                       &        1110.36         &        86.81         &        $2^{(+)}$,$3^{(-)}$          &          1965.38         &        203.34         &        $1^{(-)}$,$3^{(-)}$          &          \textbf{143.28}         &        54.20         &        $1^{(+)}$,$2^{(+)}$       \\
        & 100                   & 500                          & 15000                      &        732.32         &        81.25         &        $2^{(+)}$,$3^{(-)}$          &          953.22         &        125.64         &        $1^{(-)}$,$3^{(-)}$          &          \textbf{45.26 }        &        13.87         &        $1^{(+)}$,$2^{(+)}$       \\ \hline
unc-s-w & 100                   & 500                          & 100                        &        \textbf{1686.42}         &        272.35         &        $2^{(+)}$,$3^{(+)}$          &          4739.46         &        1283.37         &        $1^{(-)}$,$3^{(-)}$          &          2693.60         &        580.74         &        $1^{(-)}$,$2^{(+)}$       \\
        & 100                   & 500                          & 1000                       &        \textbf{262.12}         &        57.43         &        $2^{(+)}$          &          766.41         &        438.22         &        $1^{(-)}$,$3^{(-)}$          &          304.96         &        124.57         &        $2^{(+)}$       \\
        & 100                   & 500                          & 5000                       &        75.09         &        16.18         &        $3^{(-)}$          &          86.91         &        42.32         &        $3^{(-)}$          &          \textbf{47.31}         &        14.83         &        $1^{(+)}$,$2^{(+)}$       \\
        & 100                   & 500                          & 15000                      &        37.60         &        10.96         &        $2^{(-)}$,$3^{(-)}$          &          28.57         &        9.70         &        $1^{(+)}$,$3^{(-)}$          &          \textbf{15.82}         &        4.18         &        $1^{(+)}$,$2^{(+)}$       \\ \hline
bou-s-c & 100                   & 500                          & 100                        &        2523.48         &        244.20         &        $2^{(+)}$,$3^{(-)}$          &          4778.00         &        498.80         &        $1^{(-)}$,$3^{(-)}$          &          \textbf{2248.91}         &        85.01         &        $1^{(+)}$,$2^{(+)}$       \\
        & 100                   & 500                          & 1000                       &        1075.70         &        144.73         &        $2^{(+)}$,$3^{(-)}$          &          1862.45         &        236.14         &        $1^{(-)}$,$3^{(-)}$          &          \textbf{343.62}         &        72.49         &        $1^{(+)}$,$2^{(+)}$       \\
        & 100                   & 500                          & 5000                       &        579.38         &        81.32         &        $2^{(+)}$,$3^{(-)}$          &          717.55         &        50.92         &        $1^{(-)}$,$3^{(-)}$          &          \textbf{99.07}         &        42.41         &        $1^{(+)}$,$2^{(+)}$       \\
        & 100                   & 500                          & 15000                      &        407.41         &        53.79         &        $3^{(-)}$          &          358.09         &        44.40         &        $3^{(-)}$          &          \textbf{33.33}         &        13.59         &        $1^{(+)}$,$2^{(+)}$       \\               
\end{tabular}
\end{table}
The most notable results occur in the case with uncorrelated similar weights and small $\sigma$. (1+1)~EA outperforms both other algorithms in this instance. This happens because of the value of $\delta$ and the weights of the instances. $\delta$ is set to $2\sigma$ in the multi-objective approaches and the weights of items are integers in $[1001, 1010]$ in this type of instance. (1+1)~EA is able to freely get closer to the optimal solutions from both directions, while the multi-objective approaches are only allowed to consider solutions in range of $[C-\delta, C+\delta]$. In other words, it is possible that there is only one solution in that range or even no solution. Thus, the multi-objective approaches either do not find any feasible solution and get penalty in the offline error, or are not able to improve their  feasible solution. Hence, multi-objective approaches have no advantage in this type of instances according to the value of $\delta$ and weights of the items, and in fact, may have a disadvantage.

On the other hand, increasing $\sigma$ to $500$ is enough for \EA and \EAd to benefit from the population again. Although (1+1)~EA outperforms them in $\tau = 100$, it is not the absolute dominant algorithm in instances with uncorrelated similar weights anymore. More precisely, the results show that there is no significant difference between their performances in bounded strongly correlated instances and $\tau = 15000$. Furthermore, the use of population in \EA even cause a significantly better performance in low frequent changes and uncorrelated similar weights instances.

\section{NSGA-II and SPEA2}\label{sec:NSGAII-SPEA2}

The results presented in Section \ref{subec:baseline-results} demonstrate the advantage of solving dynamic KP as a multi-objective optimization problem. We showed that by using populations, specifically in low frequency changes, we can improve the efficiency and the quality of found solutions such as the algorithms are able to deal with the coming dynamic changes. In this section, we analyze the performance of NSGA-II and SPEA2 on the dynamic knapsack problem.

Both algorithms have become a well-established presence in the area of evolutionary multi-objective optimization. We are interested in analyzing the advantages of their heuristic techniques in comparison to each other and also in the environments with high frequency changes. Moreover, we compare their performance with \EAd as the best baseline algorithm of the previous section. 
\subsection{Algorithms}

\begin{algorithm}[t]
	Update $C$\;
	Update objective values of solutions in population set $P_t$ and offspring set $Q_t$\;
	\While{stopping criterion not met}{
		$R_t \leftarrow P_t\cup Q_t$
		\Comment{combine parent and offspring population}\\
		$\mathcal{F} \leftarrow$ non-dominated-sort($R_t$)
		\\
		\Comment{$\mathcal{F}=(\mathcal{F}_1,\mathcal{F}_1,\cdots)$, all non-dominated fronts of $R_t$}\\
		$P_{t+1}\leftarrow \emptyset$ and $i\leftarrow 1$\;
		\While{$|P_{t+1}|+|\mathcal{F}_i|\leq N$}{
			$P_{t+1} \leftarrow P_{t+1} \cup \mathcal{F}_i$\;
			$i\leftarrow i+1$\;
		}
		crowding-distance-assignment($\mathcal{F}_i$)\;
		Sort $\mathcal{F}_i$ based on the crowding distance in descending order\;
		$P_{t+1} \leftarrow P_{t+1}\cup \mathcal{F}_i[1:(N-|P_{t+1}|)]$\;
		\label{lin:elit-nsga}$Q_{t+1}\leftarrow$ make-new-pop($P_{t+1})$\;
		$t\leftarrow t+1$\;
	}
	\caption{NSGA-II}\label{alg:nsgaii}
\end{algorithm}
In the initial state, NSGA-II starts with a randomly generated population $P_0$ and assigns a fitness (or rank) to each solution based on its non-dominated rank. Offspring population $Q_0$ is then produced by using selection, recombination, and mutation operators. Algorithm \ref{alg:nsgaii} describes how NSGA-II performs after the initial step and after each dynamic change. It sorts the combination of offspring and population sets into a set of non-dominated fronts $\mathcal{F}=\{\mathcal{F}_1,\mathcal{F}_2,\cdots\}$ such that solutions in $\mathcal{F}_1$ are non-dominated solutions, solutions in $\mathcal{F}_2$ are non-dominated solutions after removing solutions of $\mathcal{F}_1$ from the combined set and so on. Then, starting from the first front, it adds solutions to $P_{t+1}$ until $i$th front which adding $\mathcal{F}_i$ exceeds the population size $N$. NSGA-II calculates crowding distance for each of the solutions to distinguish solutions in the same fronts. The crowding distance is estimation of perimeter of cuboid around the solution formed by using the nearest neighbors as vertices. Thus, larger crowding distance means that the solution is in more sparse area within the solutions in the same front. It assigns infinity value to the solutions with the highest or lowest objective values in each front. The algorithm finally produces $Q_{t+1}$ from $P_{t+1}$ by using evolutionary operators and considering the front rank and crowding distance as the objectives.

SPEA2, on the other hand, applies different approaches to produce distributed solutions. This algorithm keeps the best non-dominated solutions of each generation on the archive set $\overline{P_{t}}$ with size $\overline{N}$ and generates population set $P_{t+1}$, with size $N$ by performing evolutionary operators on $\overline{P_{t}}$. The fitness value of solution $x$ in SPEA2 is calculated based on two factors: integer raw fitness $0\leq R(x)$ which represent the non-dominance power of solutions that dominate $x$, and density estimate $0<D(x)\leq 1/2$ which is calculated based on the inverse of distance to the $k$th nearest neighbor of $x$ with the same raw fitness value. $k = \sqrt{N+\overline{N}}$ is chosen as the default value of $k$. The final fitness value of $x$ is calculated as $F(x) =R(x)+D(x)$. Note that $R(x)$ is $0$ when $x$ is a non-dominated solution, and the smaller value of $D(x)$ illustrates the better distributed solution.

\begin{algorithm}[t]
	Update $C$\;
	Update objective values of solutions in populations set $P_t$ and archive set $\overline{P_t}$\;
	\While{stopping criterion not met}{
		\textbf{Mating selection:} Generate a mating pool by tournament selection from $\overline{P_t}$\;
		\textbf{Variation: }Apply crossover and mutation operators on the mating pool to produce $P_{t+1}$\;
		\textbf{Fitness assignment: }Calculate fitness values of solutions in $P_{t+1} \cup \overline{P_{t}}$\;
		\label{lin:enviro-selec}\textbf{Environmental selection: } Generate $\overline{P_{t+1}}$ by choosing $\overline{N}$ non-dominated solutions from $P_{t+1} \cup\overline{P_{t}}$\;
		
	}
	\caption{SPEA2}\label{alg:spea2}
\end{algorithm}
As both of these algorithms, unlike \EAd, use specific techniques to guarantee a well-distributed solution set, to consider the capacity constraint and also prepare the algorithm for the following dynamic changes, we present a new formulation of the problem. Moreover, we discuss the elitism in each of the algorithms. We show plain elitism, which favors distributed solutions, causes the loss of good quality solution with respect to the profit and the capacity constraint. Next, we apply an additional elitism to improve the performance by keeping the best solutions in the population. 
\subsection{New Formulation for Dynamic KP}

In this section, we present a new fitness evaluation approach, different from the one used for \EAd, for NSGA-II and SPEA2 to solve the dynamic knapsack problem. In contrast to \EAd which uses two separate solution sets to store infeasible solutions to prepare for the following dynamic changes, we benefit from the ability of SPEA2 and NSGA-II in finding a well-distributed non-dominated set. We force the algorithms to find non-dominated solutions with weights within the interval $[C-\delta,C+\delta]$. To this aim, we apply penalty on the weights and profits of solutions outside of the interval such that for any solution $x$ we have
\begin{equation}
w_{\tiny MO}(x)=
\begin{cases}
w(x)   & \text{if}\ w(x)\in [C-\delta,C+\delta] \\
w(x)+(n\cdot w_{\max}+1)  \cdot \alpha(x)   & \text{otherwise,}
\end{cases}
\end{equation}
where for solution $x$ that $w(x)\notin [C-\delta,C+\delta]$, $\alpha(x) = \min\{|w(x)-C-\delta|,|w(x)-C+\delta|\}$ is the distance from the edge of the interval.
Similar to the weight, we apply penalty on the profit as follows
\begin{equation}
p_{\tiny MO}(x)=
\begin{cases}
p(x)   & \text{if}\ w(x)\in [C-\delta,C+\delta] \\
p(x)-(n\cdot p_{\max}+1)  \cdot \alpha(x)   & \text{otherwise.}
\end{cases}
\end{equation}
Note that the objectives are weight \big($w_{MO}(x)$\big) and profit \big($p_{MO}(x)$\big) which should be minimized and maximized, respectively. The penalty guarantees that any solution in the preferred interval dominates the solutions outside and solutions that are closer to the interval dominate farther ones. In this way if the algorithms produce a well-distributed non-dominated set of solutions, we expect to have good quality feasible solutions even after the dynamic change.
\subsection{Additional Elitism}
\label{sec:Additional-elitism}

\begin{figure}[t]
	\centering
	\includegraphics[width=0.5\textwidth]{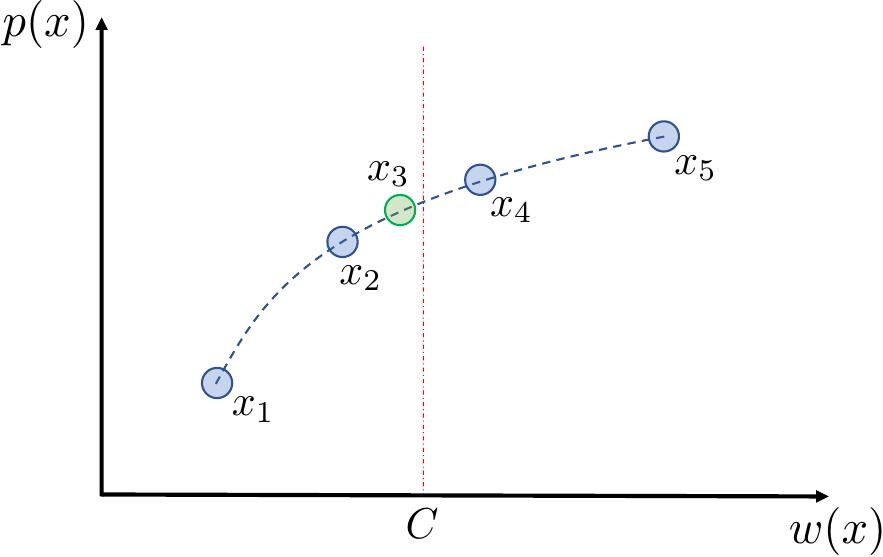}
	\caption[]
	{\centering\small A situation that NSGA-II and SPEA2 lose the best feasible solution because of the impact of distance assignment on the solution ranking.} 
	\label{fig:crowdingdist}
\end{figure}

While we still look for the feasible solution with the highest profit, the new formulation does not apply penalties on all infeasible solutions and there are better solutions in the population in terms of profit value while their weight exceed the capacity constraint. 
In NSGA-II and SPEA2, solutions are selected based on dominance ranking and crowding or density distance assignments.
Although the best feasible solution is a non-dominated solution, it is possible that the algorithms lose it because of the distance assignment. Figure \ref{fig:crowdingdist} demonstrates such a situation in which the algorithms have to pick four solutions from non-dominated set $\{x_1,\cdots,x_5\}$ to produce the next generation. Note that solutions $x_1$, $x_2$ and $x_3$ are feasible, and the others are infeasible. Both of the algorithms tend to keep $x_1$ and $x_5$ since they are border solutions and have the maximum distance value. Thus, they have to choose two solutions within $\{x_2,x_3,x_4\}$. Although, $x_3$ is the best feasible solution in this situation, the algorithms pick $x_2$ and $x_4$ since they provide a better distributed set. To address this problem, we artificially change the distance value of the best feasible solution so that the algorithms keep it in the next generation. 

To apply it on the NSGA-II, we store the best feasible solution in a separate variable. After line \ref{lin:elit-nsga}, we check if a new feasible solution dominated the previous one. If the stored best solution has been removed from the population set after line \ref{lin:elit-nsga}, and the best feasible solution in $P_{t+1}$ has less profit value, we artificially remove the worst solution from $P_{t+1}$ with regards to the front ranks and crowding distances, add the stored best solution to the first front of $P_{t+1}$, and assign infinity value to its crowding distance. Otherwise, either the best feasible solution is still in $P_{t+1}$ or the algorithm has found a better feasible solution. Hence, we only update the stored best feasible solution and assign the infinity value to its crowding distance. Note that by performing this approach, we assign more reproduction power to the best feasible solution since it is always the winner of selection phase. Hence, it is more probable that we update the feasible solution close to the constraint and achieve a better feasible solution.

In SPEA2, the elitism procedure is similar to NSGA-II. We store the best feasible solution and either update it or artificially add it to archive set in each generation after environmental selection (Line \ref{lin:enviro-selec} in Algorithm \ref{alg:spea2}). However, to make sure that it has a higher probability of reproduction, we assign zero to its fitness value. In this case, the solution is the outcome of tournament selection phase and has more opportunity to produce an offspring.

Note that changing the fitness values happens exactly before the reproduction steps. Since both algorithms re-evaluate fitness values prior to selecting the parents, the elitism approach guarantees that if another solution dominates the current best feasible solution then we remove it from the population. Hence, it does not affect the natural behavior of the algorithms. 
\subsection{Experimental Results}

We compare NSGA-II and SPEA2 results with \EAd as the previous winner algorithm in most of the cases. While the definition of dominance is the only limitation on the population size in \EAd, we set the population size for NSGA-II and SPEA2 to 20.

In addition to the total offline error, introduced in Section \ref{subsec: bech and exp}, we also compare the algorithms based on a new factor, called partial offline error. This factor considers the best feasible solution achieved by an algorithm right before a dynamic change, i.e., the performance of algorithms are analyzed based on their final population only. This factor illustrates the final achievement of the algorithms during a "no change" period and does not consider the performance of the algorithms within the optimization process. To compute the best partial offline error, instead of all the generations, we record the profit value of best feasible solution only in the last generation before the next change. Let $x_i$ denote the best feasible solution before the $(i+1)$-st change happens and $x^*_i$ denote the optimal solution corresponding to those capacity. If there is no feasible solution in the last generation, $y_i$ denotes the solution with lowest $\nu(x)$, the constraint violation of solution $x$. Then, the partial offline error for $10^6$ generations where a dynamic change takes place every $\tau$ generation is calculated as follow:
\begin{equation}
E_{\tiny BO} =\sum_{i=1}^{\lfloor10^6/\tau\rfloor}{ \frac{e_i}{\lfloor10^6/\tau\rfloor}}\text{,}
\end{equation}
where $e(x)$ if the offline error of solution $x$
calculated as follows:
\begin{equation}
e_i=
\begin{cases}
p(x^*_i) - p(x_i)   & \text{if}\ \nu(x_i) = 0 \\
p(x^*_i) + \nu(y_i)   & \text{otherwise.}
\end{cases}
\end{equation}

\begin{sidewaystable}
	\centering
	\tiny
	\setlength{\tabcolsep}{1.8pt}
	
	\caption{The mean, standard deviation values and statistical tests of the total offline error for \EAd, NSGA-II, SPEA2, NSGA-II with elitism and, SPEA2 with elitism based on the uniform distribution.}
	\label{tbl:tot_unif}
	\begin{tabular}{llllrrlrrlrrlrrlrrl}
		& \multicolumn{1}{c}{$n$} & \multicolumn{1}{c}{$r$} & \multicolumn{1}{c}{$\tau$} & \multicolumn{3}{c}{\EAd (1)}     & \multicolumn{3}{c}{NSGA-II (2)}                                                 & \multicolumn{3}{c}{SPEA2 (3)}    & \multicolumn{3}{c}{NSGA-II(we) (4)}         & \multicolumn{3}{c}{SPEA2(we) (5)}                                                                \\
		&                            &                       &                            & \multicolumn{1}{c}{mean} & \multicolumn{1}{c}{st} & \multicolumn{1}{c}{stat} & \multicolumn{1}{c}{mean} & \multicolumn{1}{c}{st} & \multicolumn{1}{c}{stat} & \multicolumn{1}{c}{mean} & \multicolumn{1}{c}{st} & \multicolumn{1}{c}{stat} & \multicolumn{1}{c}{mean} & \multicolumn{1}{c}{st} & \multicolumn{1}{c}{stat} &
		\multicolumn{1}{c}{mean} &
		\multicolumn{1}{c}{st} & \multicolumn{1}{c}{stat}  \\ \hline
		uncor   & 100                        & 2000                  & 100                        &        3684.26         &        525.50         &        $2^{(-)}$,$3^{(-)}$,$4^{(-)}$,$5^{(-)}$          &          134.29         &        30.53         &        $1^{(+)}$,$4^{(-)}$,$5^{(-)}$          &          123.12         &        31.65         &        $1^{(+)}$,$4^{(-)}$,$5^{(-)}$          &          \textbf{48.72}         &        14.43         &        $1^{(+)}$,$2^{(+)}$,$3^{(+)}$         &          50.74         &        18.50         &        $1^{(+)}$,$2^{(+)}$,$3^{(+)}$      \\
		& 100                        & 2000                  & 1000                       &        776.14         &        334.69         &        $2^{(-)}$,$3^{(-)}$,$4^{(-)}$,$5^{(-)}$          &          126.57         &        66.70         &        $1^{(+)}$,$4^{(-)}$,$5^{(-)}$          &          103.17         &        57.02         &        $1^{(+)}$,$4^{(-)}$,$5^{(-)}$          &          8.05         &        4.76         &        $1^{(+)}$,$2^{(+)}$,$3^{(+)}$         &          \textbf{6.51}         &        5.40         &        $1^{(+)}$,$2^{(+)}$,$3^{(+)}$      \\
		& 100                        & 2000                  & 5000                       &        270.90         &        121.44         &        $4^{(-)}$,$5^{(-)}$          &          168.85         &        112.04         &        $4^{(-)}$,$5^{(-)}$          &          137.66         &        92.84         &        $4^{(-)}$,$5^{(-)}$          &          3.05         &        2.17         &        $1^{(+)}$,$2^{(+)}$,$3^{(+)}$         &          \textbf{2.33}         &        3.26         &        $1^{(+)}$,$2^{(+)}$,$3^{(+)}$      \\
		& 100                        & 2000                  & 15000                      &        88.80         &        43.98         &        $4^{(-)}$,$5^{(-)}$          &          140.99         &        141.45         &        $4^{(-)}$,$5^{(-)}$          &          115.72         &        116.15         &        $4^{(-)}$,$5^{(-)}$          &          1.04         &        1.20         &        $1^{(+)}$,$2^{(+)}$,$3^{(+)}$         &          \textbf{0.62}        &        1.03         &        $1^{(+)}$,$2^{(+)}$,$3^{(+)}$      \\ \hline
		unc-s-w & 100                        & 2000                  & 100                        &        2106.45         &        249.28         &        $2^{(-)}$,$3^{(-)}$,$4^{(-)}$,$5^{(-)}$          &          10.35         &        2.43         &        $1^{(+)}$,$3^{(+)}$,$5^{(+)}$          &          134.24         &        40.80         &        $1^{(+)}$,$2^{(-)}$,$4^{(-)}$         &          \textbf{8.71}         &        1.92         &        $1^{(+)}$,$3^{(+)}$,$5^{(+)}$          &          64.11         &        20.40         &        $1^{(+)}$,$2^{(-)}$,$4^{(-)}$      \\
		& 100                        & 2000                  & 1000                       &        302.34         &        24.60         &        $2^{(-)}$,$4^{(-)}$,$5^{(-)}$          &          3.37         &        2.35         &        $1^{(+)}$,$3^{(+)}$,$5^{(+)}$          &          209.79         &        96.26         &        $2^{(-)}$,$4^{(-)}$,$5^{(-)}$          &          \textbf{1.18}         &        0.51         &        $1^{(+)}$,$3^{(+)}$,$5^{(+)}$          &          24.76         &        15.37         &        $1^{(+)}$,$2^{(-)}$,$3^{(+)}$,$4^{(-)}$      \\
		& 100                        & 2000                  & 5000                       &        60.94         &        9.12         &        $2^{(-)}$,$4^{(-)}$,$5^{(-)}$          &          3.78         &        3.02         &        $1^{(+)}$,$3^{(+)}$         &          357.82         &        125.60         &        $2^{(-)}$,$4^{(-)}$,$5^{(-)}$          &          \textbf{0.56}         &        0.39         &        $1^{(+)}$,$3^{(+)}$,$5^{(+)}$          &          13.42         &        5.70         &        $1^{(+)}$,$3^{(+)}$,$4^{(-)}$      \\
		& 100                        & 2000                  & 15000                      &        19.26         &        4.04         &        $2^{(-)}$,$4^{(-)}$,$5^{(-)}$          &          4.08         &        4.43         &        $1^{(+)}$,$3^{(+)}$         &          353.30         &        166.59         &        $2^{(-)}$,$4^{(-)}$,$5^{(-)}$          &          \textbf{0.36}         &        0.47         &        $1^{(+)}$,$3^{(+)}$,$5^{(+)}$          &          5.29         &        4.21         &        $1^{(+)}$,$3^{(+)}$,$4^{(-)}$      \\ \hline
		bou-s-c & 100                        & 2000                  & 100                        &        3036.97         &        297.34         &        $3^{(-)}$,$4^{(-)}$,$5^{(-)}$          &          224.80         &        6.82         &        $4^{(-)}$,$5^{(-)}$          &          203.96         &        6.62         &        $1^{(+)}$,$5^{(-)}$          &          71.96         &        2.46         &        $1^{(+)}$,$2^{(+)}$         &          \textbf{58.13}         &        4.74         &        $1^{(+)}$,$2^{(+)}$,$3^{(+)}$      \\
		& 100                        & 2000                  & 1000                       &        617.92         &        186.35         &        $3^{(-)}$,$4^{(-)}$,$5^{(-)}$          &          235.49         &        15.10         &        $4^{(-)}$,$5^{(-)}$          &          200.86         &        9.40         &        $1^{(+)}$,$5^{(-)}$          &          24.82         &        1.96         &        $1^{(+)}$,$2^{(+)}$         &          \textbf{15.36}         &        1.61         &        $1^{(+)}$,$2^{(+)}$,$3^{(+)}$      \\
		& 100                        & 2000                  & 5000                       &        251.41         &        109.58         &        $4^{(-)}$,$5^{(-)}$          &          242.63         &        17.76         &        $3^{(-)}$,$4^{(-)}$,$5^{(-)}$          &          204.60         &        7.40         &        $2^{(+)}$,$4^{(-)}$,$5^{(-)}$          &          10.36         &        2.44         &        $1^{(+)}$,$2^{(+)}$,$3^{(+)}$         &          \textbf{4.76}         &        1.21         &        $1^{(+)}$,$2^{(+)}$,$3^{(+)}$      \\
		& 100                        & 2000                  & 15000                      &        104.27         &        61.06         &        $2^{(+)}$,$4^{(-)}$,$5^{(-)}$          &          244.34         &        25.98         &        $1^{(-)}$,$4^{(-)}$,$5^{(-)}$          &          207.38         &        12.74         &        $4^{(-)}$,$5^{(-)}$          &          5.74         &        2.27         &        $1^{(+)}$,$2^{(+)}$,$3^{(+)}$         &          \textbf{2.07}         &        1.08         &        $1^{(+)}$,$2^{(+)}$,$3^{(+)}$      \\ \hline
		uncor   & 100                        & 10000                 & 100                        &        11038.07         &        236.91         &        $2^{(-)}$,$4^{(-)}$,$5^{(-)}$          &          1227.93         &        62.11         &        $1^{(+)}$,$3^{(+)}$,$4^{(-)}$         &          1688.41         &        95.98         &        $2^{(-)}$,$4^{(-)}$,$5^{(-)}$          &          \textbf{944.55}         &        57.23         &        $1^{(+)}$,$2^{(+)}$,$3^{(+)}$,$5^{(+)}$          &          1387.19         &        92.72         &        $1^{(+)}$,$3^{(+)}$,$4^{(-)}$      \\
		& 100                        & 10000                 & 1000                       &        3508.51         &        473.42         &        $2^{(-)}$,$3^{(-)}$,$4^{(-)}$,$5^{(-)}$          &          898.80         &        126.18         &        $1^{(+)}$,$4^{(-)}$,$5^{(-)}$          &          1083.25         &        203.47         &        $1^{(+)}$,$4^{(-)}$,$5^{(-)}$          &          \textbf{197.58}        &        34.04         &        $1^{(+)}$,$2^{(+)}$,$3^{(+)}$         &          380.65         &        94.12         &        $1^{(+)}$,$2^{(+)}$,$3^{(+)}$      \\
		& 100                        & 10000                 & 5000                       &        1183.52         &        411.83         &        $4^{(-)}$,$5^{(-)}$          &          951.97         &        256.55         &        $4^{(-)}$,$5^{(-)}$          &          987.39         &        351.94         &        $4^{(-)}$,$5^{(-)}$          &          \textbf{99.76}         &        32.32         &        $1^{(+)}$,$2^{(+)}$,$3^{(+)}$         &          106.14         &        50.70         &        $1^{(+)}$,$2^{(+)}$,$3^{(+)}$      \\
		& 100                        & 10000                 & 15000                      &        566.70         &        219.54         &        $4^{(-)}$,$5^{(-)}$          &          953.69         &        459.12         &        $4^{(-)}$,$5^{(-)}$          &          927.86         &        613.20         &        $4^{(-)}$,$5^{(-)}$          &          59.96         &        35.22         &        $1^{(+)}$,$2^{(+)}$,$3^{(+)}$         &          \textbf{40.05}        &        32.27         &        $1^{(+)}$,$2^{(+)}$,$3^{(+)}$      \\ \hline
		unc-s-w & 100                        & 10000                 & 100                        &        8057.44         &        274.17         &        $2^{(-)}$,$4^{(-)}$,$5^{(-)}$          &          562.54         &        34.34         &        $1^{(+)}$,$4^{(-)}$         &          653.70         &        41.75         &        $4^{(-)}$,$5^{(-)}$          &          \textbf{283.83}        &        18.20         &        $1^{(+)}$,$2^{(+)}$,$3^{(+)}$         &          396.46         &        34.89         &        $1^{(+)}$,$3^{(+)}$      \\
		& 100                        & 10000                 & 1000                        &        1743.12         &        364.38         &        $2^{(-)}$,$3^{(-)}$,$4^{(-)}$,$5^{(-)}$          &          601.19         &        108.71         &        $1^{(+)}$,$4^{(-)}$,$5^{(-)}$          &          584.00         &        106.60         &        $1^{(+)}$,$4^{(-)}$,$5^{(-)}$          &          116.02         &        26.00         &        $1^{(+)}$,$2^{(+)}$,$3^{(+)}$         &          \textbf{107.31}         &        28.42         &        $1^{(+)}$,$2^{(+)}$,$3^{(+)}$      \\
		& 100                        & 10000                 & 5000                       &        519.63         &        175.22         &        $4^{(-)}$,$5^{(-)}$          &          638.72         &        224.53         &        $4^{(-)}$,$5^{(-)}$          &          593.92         &        226.11         &        $4^{(-)}$,$5^{(-)}$          &          58.83         &        30.19         &        $1^{(+)}$,$2^{(+)}$,$3^{(+)}$         &          \textbf{41.63}         &        24.35         &        $1^{(+)}$,$2^{(+)}$,$3^{(+)}$      \\
		& 100                        & 10000                 & 15000                      &        201.97         &        79.28         &        $2^{(+)}$,$3^{(+)}$,$4^{(-)}$,$5^{(-)}$          &          618.73         &        277.40         &        $1^{(-)}$,$4^{(-)}$,$5^{(-)}$          &          549.70         &        266.72         &        $1^{(-)}$,$4^{(-)}$,$5^{(-)}$          &          25.41         &        19.56         &        $1^{(+)}$,$2^{(+)}$,$3^{(+)}$         &          \textbf{15.05}         &        13.88         &        $1^{(+)}$,$2^{(+)}$,$3^{(+)}$      \\ \hline
		bou-s-c & 100                        & 10000                 & 100                        &        12736.55         &        229.48         &        $2^{(-)}$,$4^{(-)}$,$5^{(-)}$          &          1449.09         &        39.45         &        $1^{(+)}$,$4^{(-)}$         &          1625.39         &        64.99         &        $4^{(-)}$,$5^{(-)}$          &          \textbf{693.73}         &        38.56         &        $1^{(+)}$,$2^{(+)}$,$3^{(+)}$         &          878.51         &        62.47         &        $1^{(+)}$,$3^{(+)}$      \\
		& 100                        & 10000                 & 1000                       &        3575.26         &        550.54         &        $2^{(-)}$,$3^{(-)}$,$4^{(-)}$,$5^{(-)}$          &          1421.59         &        105.78         &        $1^{(+)}$,$4^{(-)}$,$5^{(-)}$          &          1515.04         &        145.19         &        $1^{(+)}$,$4^{(-)}$,$5^{(-)}$          &          \textbf{306.22}         &        38.62         &        $1^{(+)}$,$2^{(+)}$,$3^{(+)}$         &          386.50         &        61.47         &        $1^{(+)}$,$2^{(+)}$,$3^{(+)}$      \\
		& 100                        & 10000                 & 5000                       &        1472.19         &        493.88         &        $4^{(-)}$,$5^{(-)}$          &          1419.66         &        223.08         &        $4^{(-)}$,$5^{(-)}$          &          1491.40         &        281.66         &        $4^{(-)}$,$5^{(-)}$          &          \textbf{176.12}         &        47.55         &        $1^{(+)}$,$2^{(+)}$,$3^{(+)}$         &          202.88         &        48.35         &        $1^{(+)}$,$2^{(+)}$,$3^{(+)}$      \\
		& 100                        & 10000                 & 15000                      &        977.42         &        397.75         &        $4^{(-)}$,$5^{(-)}$          &          1349.77         &        294.06         &        $4^{(-)}$,$5^{(-)}$          &          1448.12         &        369.69         &        $4^{(-)}$,$5^{(-)}$          &          \textbf{109.48}         &        40.90         &        $1^{(+)}$,$2^{(+)}$,$3^{(+)}$         &          129.12         &        41.19         &        $1^{(+)}$,$2^{(+)}$,$3^{(+)}$                           
	\end{tabular}
\end{sidewaystable}



\begin{sidewaystable}

	\setlength{\tabcolsep}{1.8pt}
	\centering
	\tiny
	\caption{The mean, standard deviation values and statistical tests of the total offline error for \EAd, NSGA-II, SPEA2, NSGA-II with elitism, and SPEA2 with elitism based on the normal distribution.}
	\label{tbl:tot_norm}
	\begin{tabular}{llllrrlrrlrrlrrlrrl}
		& \multicolumn{1}{c}{$n$} & \multicolumn{1}{c}{$\sigma$} & \multicolumn{1}{c}{$\tau$} & \multicolumn{3}{c}{\EAd (1)}     & \multicolumn{3}{c}{NSGA-II (2)}                                                 & \multicolumn{3}{c}{SPEA2 (3)}    & \multicolumn{3}{c}{NSGA-II(we) (4)}         & \multicolumn{3}{c}{SPEA2(we) (5)}                                                                \\
		&                            &                       &                            & \multicolumn{1}{c}{mean} & \multicolumn{1}{c}{st} & \multicolumn{1}{c}{stat} & \multicolumn{1}{c}{mean} & \multicolumn{1}{c}{st} & \multicolumn{1}{c}{stat} & \multicolumn{1}{c}{mean} & \multicolumn{1}{c}{st} & \multicolumn{1}{c}{stat} & \multicolumn{1}{c}{mean} & \multicolumn{1}{c}{st} & \multicolumn{1}{c}{stat} &
		\multicolumn{1}{c}{mean} &
		\multicolumn{1}{c}{st} & \multicolumn{1}{c}{stat}  \\ \hline
		uncor   & 100                   & 100                          & 100                        &        4271.09         &        789.94         &        $2^{(-)}$,$3^{(-)}$,$4^{(-)}$,$5^{(-)}$          &          9.67         &        1.85         &        $1^{(+)}$,$3^{(-)}$,$5^{(-)}$          &          6.66         &        2.25         &        $1^{(+)}$,$2^{(+)}$         &          7.34         &        2.08         &        $1^{(+)}$         &          \textbf{5.48}         &        2.59         &        $1^{(+)}$,$2^{(+)}$      \\
		& 100                   & 100                          & 1000                      &        412.89         &        27.25         &        $2^{(-)}$,$3^{(-)}$,$4^{(-)}$,$5^{(-)}$          &          3.79         &        1.89         &        $1^{(+)}$,$3^{(-)}$,$4^{(-)}$,$5^{(-)}$          &          1.61         &        1.14         &        $1^{(+)}$,$2^{(+)}$,$5^{(-)}$          &          1.08         &        0.45         &        $1^{(+)}$,$2^{(+)}$         &          \textbf{0.58}         &        0.31         &        $1^{(+)}$,$2^{(+)}$,$3^{(+)}$      \\
		& 100                   & 100                          & 5000                       &        108.29         &        14.22         &        $2^{(-)}$,$3^{(-)}$,$4^{(-)}$,$5^{(-)}$          &          5.03         &        3.59         &        $1^{(+)}$,$4^{(-)}$,$5^{(-)}$          &          2.82         &        2.95         &        $1^{(+)}$,$5^{(-)}$          &          0.44         &        0.34         &        $1^{(+)}$,$2^{(+)}$         &          \textbf{0.11}         &        0.10         &        $1^{(+)}$,$2^{(+)}$,$3^{(+)}$      \\
		& 100                   & 100                          & 15000                      &        61.93         &        17.03         &        $2^{(-)}$,$3^{(-)}$,$4^{(-)}$,$5^{(-)}$          &          3.79         &        2.79         &        $1^{(+)}$,$4^{(-)}$,$5^{(-)}$          &          1.82         &        2.12         &        $1^{(+)}$,$5^{(-)}$          &          0.18         &        0.18         &        $1^{(+)}$,$2^{(+)}$         &          \textbf{0.05}         &        0.08         &        $1^{(+)}$,$2^{(+)}$,$3^{(+)}$      \\ \hline
		unc-s-w & 100                   & 100                          & 100                        &        \textbf{1904.09}         &        877.55         &        $2^{(+)}$,$3^{(+)}$         &          7875.64         &        4651.35         &        $1^{(-)}$,$4^{(-)}$,$5^{(-)}$          &          7755.18         &        4609.32         &        $1^{(-)}$,$4^{(-)}$,$5^{(-)}$          &          3905.34         &        2496.65         &        $2^{(+)}$,$3^{(+)}$         &          3526.70         &        2150.96         &        $2^{(+)}$,$3^{(+)}$      \\
		& 100                   & 100                          & 1000                       &        \textbf{1482.37}         &        391.75         &        $2^{(+)}$,$3^{(+)}$,$4^{(+)}$,$5^{(+)}$          &          8914.58         &        1233.57         &        $1^{(-)}$,$4^{(-)}$,$5^{(-)}$          &          8981.25         &        1260.77         &        $1^{(-)}$,$4^{(-)}$,$5^{(-)}$          &          4264.04         &        1011.79         &        $1^{(-)}$,$2^{(+)}$,$3^{(+)}$         &          4169.18         &        992.44         &        $1^{(-)}$,$2^{(+)}$,$3^{(+)}$      \\
		& 100                   & 100                          & 5000                       &        \textbf{1322.35}         &        414.28         &        $2^{(+)}$,$3^{(+)}$         &          4490.67         &        1477.30         &        $1^{(-)}$,$4^{(-)}$,$5^{(-)}$          &          4502.98         &        1486.96         &        $1^{(-)}$,$4^{(-)}$,$5^{(-)}$          &          2102.75         &        1241.46         &        $2^{(+)}$,$3^{(+)}$         &          2135.04         &        1202.39         &        $2^{(+)}$,$3^{(+)}$      \\
		& 100                   & 100                          & 15000                      &       \textbf{1137.80}         &        648.73         &        $2^{(+)}$,$3^{(+)}$         &          4585.86         &        1141.78         &        $1^{(-)}$,$4^{(-)}$,$5^{(-)}$          &          4598.89         &        1147.26         &        $1^{(-)}$,$4^{(-)}$,$5^{(-)}$          &          1863.60         &        1321.14         &        $2^{(+)}$,$3^{(+)}$         &          1954.11         &        1455.11         &        $2^{(+)}$,$3^{(+)}$      \\ \hline
		bou-s-c & 100                   & 100                          & 100                       &        2903.77         &        717.92         &        $2^{(-)}$,$3^{(-)}$,$4^{(-)}$,$5^{(-)}$          &          20.47         &        3.55         &        $1^{(+)}$,$3^{(-)}$,$4^{(-)}$,$5^{(-)}$          &          14.45         &        2.00         &        $1^{(+)}$,$2^{(+)}$,$5^{(-)}$          &          11.99         &        2.80         &        $1^{(+)}$,$2^{(+)}$         &          \textbf{9.80}         &        1.97         &        $1^{(+)}$,$2^{(+)}$,$3^{(+)}$      \\
		& 100                   & 100                          & 1000                       &        312.88         &        35.53         &        $3^{(-)}$,$4^{(-)}$,$5^{(-)}$          &          17.22         &        0.73         &        $4^{(-)}$,$5^{(-)}$          &          12.32         &        1.10         &        $1^{(+)}$,$4^{(-)}$,$5^{(-)}$          &          3.89         &        0.60         &        $1^{(+)}$,$2^{(+)}$,$3^{(+)}$         &          \textbf{3.61}         &        1.15         &        $1^{(+)}$,$2^{(+)}$,$3^{(+)}$      \\
		& 100                   & 100                          & 5000                       &        101.21         &        17.47         &        $3^{(-)}$,$4^{(-)}$,$5^{(-)}$          &          17.02         &        0.84         &        $4^{(-)}$,$5^{(-)}$          &          12.62         &        1.97         &        $1^{(+)}$,$4^{(-)}$,$5^{(-)}$          &          1.44         &        0.30         &        $1^{(+)}$,$2^{(+)}$,$3^{(+)}$         &          \textbf{1.10}         &        0.40         &        $1^{(+)}$,$2^{(+)}$,$3^{(+)}$      \\
		& 100                   & 100                          & 15000                      &        70.16         &        22.26         &        $2^{(-)}$,$3^{(-)}$,$4^{(-)}$,$5^{(-)}$          &          17.04         &        0.85         &        $1^{(+)}$,$4^{(-)}$,$5^{(-)}$          &          13.37         &        2.13         &        $1^{(+)}$,$4^{(-)}$,$5^{(-)}$          &          0.73         &        0.25         &        $1^{(+)}$,$2^{(+)}$,$3^{(+)}$         &          \textbf{0.46}         &        0.38         &        $1^{(+)}$,$2^{(+)}$,$3^{(+)}$      \\ \hline
		uncor   & 100                   & 500                          & 100                        &        2469.58         &        649.04         &        $2^{(-)}$,$3^{(-)}$,$4^{(-)}$,$5^{(-)}$          &          54.72         &        30.38         &        $1^{(+)}$,$4^{(-)}$,$5^{(-)}$          &          41.16         &        24.57         &        $1^{(+)}$,$4^{(-)}$,$5^{(-)}$          &          16.93         &        9.82         &        $1^{(+)}$,$2^{(+)}$,$3^{(+)}$         &          \textbf{13.81}        &        9.04         &        $1^{(+)}$,$2^{(+)}$,$3^{(+)}$      \\
		& 100                   & 500                          & 1000                       "&        511.58         &        187.21         &        $2^{(-)}$,$3^{(-)}$,$4^{(-)}$,$5^{(-)}$          &          70.79         &        52.31         &        $1^{(+)}$,$4^{(-)}$,$5^{(-)}$          &          51.23         &        39.45         &        $1^{(+)}$,$4^{(-)}$,$5^{(-)}$          &          3.75         &        2.51         &        $1^{(+)}$,$2^{(+)}$,$3^{(+)}$         &          \textbf{2.47}         &        1.87         &        $1^{(+)}$,$2^{(+)}$,$3^{(+)}$      \\
		& 100                   & 500                          & 5000                       &        143.28         &        54.20         &        $4^{(-)}$,$5^{(-)}$          &          108.83         &        57.57         &        $4^{(-)}$,$5^{(-)}$          &          80.00         &        45.64         &        $4^{(-)}$,$5^{(-)}$          &          1.35         &        0.45         &        $1^{(+)}$,$2^{(+)}$,$3^{(+)}$         &          \textbf{0.88}         &        0.63         &        $1^{(+)}$,$2^{(+)}$,$3^{(+)}$      \\
		& 100                   & 500                          & 15000                      &        45.26         &        13.87         &        $2^{(+)}$,$3^{(+)}$,$4^{(-)}$,$5^{(-)}$          &          142.80         &        60.99         &        $1^{(-)}$,$4^{(-)}$,$5^{(-)}$          &          107.71         &        48.50         &        $1^{(-)}$,$4^{(-)}$,$5^{(-)}$          &          0.58         &        0.18         &        $1^{(+)}$,$2^{(+)}$,$3^{(+)}$         &          \textbf{0.44}         &        0.38         &        $1^{(+)}$,$2^{(+)}$,$3^{(+)}$      \\ \hline
		unc-s-w & 100                   & 500                          & 100                        &        2693.60         &        580.74         &        $2^{(-)}$,$3^{(-)}$,$4^{(-)}$,$5^{(-)}$          &          29.12         &        8.72         &        $1^{(+)}$,$3^{(+)}$,$5^{(+)}$          &          164.38         &        61.26         &        $1^{(+)}$,$2^{(-)}$,$4^{(-)}$         &          \textbf{28.72}         &        8.57         &        $1^{(+)}$,$3^{(+)}$,$5^{(+)}$          &          78.35         &        29.19         &        $1^{(+)}$,$2^{(-)}$,$4^{(-)}$      \\
		& 100                   & 500                          & 1000                       &        304.96         &        124.57         &        $2^{(-)}$,$4^{(-)}$,$5^{(-)}$          &          6.92         &        2.73         &        $1^{(+)}$,$3^{(+)}$,$5^{(+)}$          &          413.96         &        221.74         &        $2^{(-)}$,$4^{(-)}$,$5^{(-)}$          &          \textbf{5.93}         &        3.00         &        $1^{(+)}$,$3^{(+)}$,$5^{(+)}$          &          52.51         &        22.42         &        $1^{(+)}$,$2^{(-)}$,$3^{(+)}$,$4^{(-)}$      \\
		& 100                   & 500                          & 5000                       &        47.31         &        14.83         &        $2^{(-)}$,$4^{(-)}$,$5^{(-)}$          &          2.65         &        1.51         &        $1^{(+)}$,$3^{(+)}$         &          402.22         &        328.16         &        $2^{(-)}$,$4^{(-)}$,$5^{(-)}$          &          \textbf{1.10}         &        1.10         &        $1^{(+)}$,$3^{(+)}$,$5^{(+)}$          &          20.05         &        16.55         &        $1^{(+)}$,$3^{(+)}$,$4^{(-)}$      \\
		& 100                   & 500                          & 15000                      &        15.82         &        4.18         &        $2^{(-)}$,$4^{(-)}$,$5^{(-)}$          &          2.26         &        2.02         &        $1^{(+)}$,$3^{(+)}$         &          264.22         &        280.33         &        $2^{(-)}$,$4^{(-)}$,$5^{(-)}$          &          \textbf{0.33}         &        0.37         &        $1^{(+)}$,$3^{(+)}$,$5^{(+)}$          &          5.62         &        6.89         &        $1^{(+)}$,$3^{(+)}$,$4^{(-)}$      \\ \hline
		bou-s-c & 100                   & 500                          & 100                        &        2248.91         &        85.01         &        $3^{(-)}$,$4^{(-)}$,$5^{(-)}$          &          98.45         &        4.84         &        $4^{(-)}$,$5^{(-)}$          &          80.20         &        3.81         &        $1^{(+)}$,$5^{(-)}$          &          32.91         &        0.98         &        $1^{(+)}$,$2^{(+)}$         &          \textbf{23.64}         &        1.43         &        $1^{(+)}$,$2^{(+)}$,$3^{(+)}$      \\
		& 100                   & 500                          & 1000                       &        343.62         &        72.49         &        $3^{(-)}$,$4^{(-)}$,$5^{(-)}$          &          106.57         &        9.43         &        $4^{(-)}$,$5^{(-)}$          &          81.93         &        6.29         &        $1^{(+)}$,$5^{(-)}$          &          9.59         &        1.48         &        $1^{(+)}$,$2^{(+)}$         &          \textbf{5.22}        &        1.50         &        $1^{(+)}$,$2^{(+)}$,$3^{(+)}$      \\
		& 100                   & 500                          & 5000                       &        99.07         &        42.41         &        $4^{(-)}$,$5^{(-)}$          &          112.97         &        10.10         &        $3^{(-)}$,$4^{(-)}$,$5^{(-)}$          &          85.87         &        8.79         &        $2^{(+)}$,$4^{(-)}$,$5^{(-)}$          &          2.82         &        0.61         &        $1^{(+)}$,$2^{(+)}$,$3^{(+)}$         &          \textbf{1.02}        &        0.26         &        $1^{(+)}$,$2^{(+)}$,$3^{(+)}$      \\
		& 100                   & 500                          & 15000                      &        33.33         &        13.59         &        $2^{(+)}$,$5^{(-)}$          &          117.62         &        11.47         &        $1^{(-)}$,$4^{(-)}$,$5^{(-)}$          &          90.26         &        10.34         &        $4^{(-)}$,$5^{(-)}$          &          1.10         &        0.24         &        $2^{(+)}$,$3^{(+)}$         &          \textbf{0.30}         &        0.08         &        $1^{(+)}$,$2^{(+)}$,$3^{(+)}$                            
	\end{tabular}

\end{sidewaystable}

\subsection{Analysis}
\label{sec:Analysis}

We now present a detailed analysis on the performance of simple and advanced baseline evolutionary algorithms on the dynamic knapsack problem in our experiments. 

In Table \ref{tbl:tot_unif}, we compare the different algorithms, namely \EAd, NSGA-II, SPEA2 and two novel algorithms, NSGA-II with new elitism, named as NSGA-II(we), and SPEA2 with new elitism named as SPEA2(we). We summarize our results in terms of the mean value of the offline error achieved for each instance based on the uniform distribution.

The displayed results show that NSGA-II and SPEA2 significantly outperform \EAd when the changes occur more frequently, i.e., $\tau = 100, 1000$ for most considered instances. However, \EAd achieves better results for the uncorrelated similar weights instances compared to NSGA-II and SPEA2 when the frequency is low. For the frequency $\tau = 15000$, \EAd especially has sufficient time to find a good solution. When comparing \EAd, NSGA-II, SPEA2 with algorithms based on elitism, NSGA-II(we) and SPEA2(we) achieve the best results among all different instances in most cases. This may indicate that both algorithms effectively used the elitism strategy outlined in Section \ref{sec:Additional-elitism}.

\begin{sidewaystable}
	\centering
	\tiny
	\setlength{\tabcolsep}{1.8pt}
	\caption{The mean, standard deviation values and statistical tests of the partial offline error for \EAd, NSGA-II, SPEA2, NSGA-II with elitism and, SPEA2 with elitism based on the uniform distribution.}
	\label{tbl:par_unif}
	
	\begin{tabular}{llllrrlrrlrrlrrlrrl}
		& \multicolumn{1}{c}{$n$} & \multicolumn{1}{c}{$r$} & \multicolumn{1}{c}{$\tau$} & \multicolumn{3}{c}{\EAd (1)}     & \multicolumn{3}{c}{NSGA-II (2)}                                                 & \multicolumn{3}{c}{SPEA2 (3)}    & \multicolumn{3}{c}{NSGA-II(we) (4)}         & \multicolumn{3}{c}{SPEA2(we) (5)}                                                                \\
		&                            &                       &                            & \multicolumn{1}{c}{mean} & \multicolumn{1}{c}{st} & \multicolumn{1}{c}{stat} & \multicolumn{1}{c}{mean} & \multicolumn{1}{c}{st} & \multicolumn{1}{c}{stat} & \multicolumn{1}{c}{mean} & \multicolumn{1}{c}{st} & \multicolumn{1}{c}{stat} & \multicolumn{1}{c}{mean} & \multicolumn{1}{c}{st} & \multicolumn{1}{c}{stat} &
		\multicolumn{1}{c}{mean} &
		\multicolumn{1}{c}{st} & \multicolumn{1}{c}{stat}  \\ \hline
		uncor   & 100                        & 2000                  & 100                        &        2134.03         &        377.28         &        $2^{(-)}$,$3^{(-)}$,$4^{(-)}$,$5^{(-)}$          &          135.95         &        29.93         &        $1^{(+)}$,$4^{(-)}$,$5^{(-)}$          &          119.59         &        29.48         &        $1^{(+)}$,$4^{(-)}$,$5^{(-)}$          &          26.04         &        8.48         &        $1^{(+)}$,$2^{(+)}$,$3^{(+)}$         &          25.86         &        10.33         &        $1^{(+)}$,$2^{(+)}$,$3^{(+)}$      \\
		& 100                        & 2000                  & 1000                       &        336.71         &        179.08         &        $3^{(-)}$,$4^{(-)}$,$5^{(-)}$          &          128.84         &        69.21         &        $4^{(-)}$,$5^{(-)}$          &          103.68         &        57.17         &        $1^{(+)}$,$4^{(-)}$,$5^{(-)}$          &          2.31         &        1.49         &        $1^{(+)}$,$2^{(+)}$,$3^{(+)}$         &          0.69         &        0.75         &        $1^{(+)}$,$2^{(+)}$,$3^{(+)}$      \\
		& 100                        & 2000                  & 5000                       &        79.80         &        47.49         &        $4^{(-)}$,$5^{(-)}$          &          168.92         &        116.40         &        $4^{(-)}$,$5^{(-)}$          &          141.67         &        99.47         &        $4^{(-)}$,$5^{(-)}$          &          0.50         &        0.47         &        $1^{(+)}$,$2^{(+)}$,$3^{(+)}$         &          0.01         &        0.01         &        $1^{(+)}$,$2^{(+)}$,$3^{(+)}$      \\
		& 100                        & 2000                  & 15000                      &        14.70         &        16.17         &        $2^{(+)}$,$3^{(+)}$,$4^{(-)}$,$5^{(-)}$          &          141.09         &        141.93         &        $1^{(-)}$,$4^{(-)}$,$5^{(-)}$          &          123.87         &        128.21         &        $1^{(-)}$,$4^{(-)}$,$5^{(-)}$          &          0.13         &        0.29         &        $1^{(+)}$,$2^{(+)}$,$3^{(+)}$         &          0.00         &        0.00         &        $1^{(+)}$,$2^{(+)}$,$3^{(+)}$      \\ \hline
		unc-s-w & 100                        & 2000                  & 100                        &        1102.09         &        113.57         &        $2^{(-)}$,$4^{(-)}$,$5^{(-)}$          &          5.71         &        1.74         &        $1^{(+)}$,$3^{(+)}$,$5^{(+)}$          &          133.25         &        40.74         &        $2^{(-)}$,$4^{(-)}$         &          3.59         &        0.90         &        $1^{(+)}$,$3^{(+)}$,$5^{(+)}$          &          54.73         &        18.74         &        $1^{(+)}$,$2^{(-)}$,$4^{(-)}$      \\       
		& 100                        & 2000                  & 1000                       &        103.15         &        15.95         &        $2^{(-)}$,$4^{(-)}$,$5^{(-)}$          &          2.55         &        2.12         &        $1^{(+)}$,$3^{(+)}$         &          233.01         &        101.48         &        $2^{(-)}$,$4^{(-)}$,$5^{(-)}$          &          0.27         &        0.27         &        $1^{(+)}$,$3^{(+)}$,$5^{(+)}$          &          17.39         &        11.62         &        $1^{(+)}$,$3^{(+)}$,$4^{(-)}$      \\
		& 100                        & 2000                  & 5000                       &        10.32         &        5.06         &        $2^{(-)}$,$3^{(+)}$,$4^{(-)}$         &          3.49         &        3.41         &        $1^{(+)}$,$3^{(+)}$,$4^{(-)}$         &          391.01         &        141.59         &        $1^{(-)}$,$2^{(-)}$,$4^{(-)}$,$5^{(-)}$          &          0.25         &        0.27         &        $1^{(+)}$,$2^{(+)}$,$3^{(+)}$,$5^{(+)}$          &          6.96         &        3.67         &        $3^{(+)}$,$4^{(-)}$      \\
		& 100                        & 2000                  & 15000                      &        1.34         &        1.05         &        $3^{(+)}$,$4^{(-)}$         &          4.27         &        4.85         &        $3^{(+)}$,$4^{(-)}$         &          381.02         &        187.18         &        $1^{(-)}$,$2^{(-)}$,$4^{(-)}$,$5^{(-)}$          &          0.21         &        0.41         &        $1^{(+)}$,$2^{(+)}$,$3^{(+)}$,$5^{(+)}$          &          1.60         &        1.94         &        $3^{(+)}$,$4^{(-)}$      \\ \hline
		bou-s-c & 100                        & 2000                  & 100                        &        1296.54         &        241.04         &        $3^{(-)}$,$4^{(-)}$,$5^{(-)}$          &          230.77         &        7.05         &        $4^{(-)}$,$5^{(-)}$          &          204.91         &        5.07         &        $1^{(+)}$,$5^{(-)}$          &          43.92         &        1.13         &        $1^{(+)}$,$2^{(+)}$         &          33.14         &        1.81         &        $1^{(+)}$,$2^{(+)}$,$3^{(+)}$      \\
		& 100                        & 2000                  & 1000                       &        259.02         &        92.31         &        $4^{(-)}$,$5^{(-)}$          &          236.93         &        16.89         &        $4^{(-)}$,$5^{(-)}$          &          201.48         &        9.66         &        $4^{(-)}$,$5^{(-)}$          &          12.84         &        1.72         &        $1^{(+)}$,$2^{(+)}$,$3^{(+)}$         &          6.00         &        1.86         &        $1^{(+)}$,$2^{(+)}$,$3^{(+)}$      \\
		& 100                        & 2000                  & 5000                       &        96.79         &        61.33         &        $2^{(+)}$,$4^{(-)}$,$5^{(-)}$          &          241.60         &        23.83         &        $1^{(-)}$,$4^{(-)}$,$5^{(-)}$          &          204.39         &        11.15         &        $4^{(-)}$,$5^{(-)}$          &          4.84         &        2.22         &        $1^{(+)}$,$2^{(+)}$,$3^{(+)}$         &          1.01         &        0.91         &        $1^{(+)}$,$2^{(+)}$,$3^{(+)}$      \\
		& 100                        & 2000                  & 15000                      &        37.60         &        33.88         &        $2^{(+)}$,$3^{(+)}$,$5^{(-)}$          &          254.73         &        34.35         &        $1^{(-)}$,$4^{(-)}$,$5^{(-)}$          &          214.77         &        19.64         &        $1^{(-)}$,$4^{(-)}$,$5^{(-)}$          &          2.58         &        1.82         &        $2^{(+)}$,$3^{(+)}$         &          0.23         &        0.41         &        $1^{(+)}$,$2^{(+)}$,$3^{(+)}$      \\ \hline
		uncor   & 100                        & 10000                 & 100                        &        9410.13         &        230.91         &        $2^{(-)}$,$4^{(-)}$,$5^{(-)}$          &          1084.42         &        58.24         &        $1^{(+)}$,$3^{(+)}$,$4^{(-)}$         &          1448.55         &        87.95         &        $2^{(-)}$,$4^{(-)}$,$5^{(-)}$          &          679.17         &        47.17         &        $1^{(+)}$,$2^{(+)}$,$3^{(+)}$,$5^{(+)}$          &          1001.91         &        77.73         &        $1^{(+)}$,$3^{(+)}$,$4^{(-)}$      \\
		& 100                        & 10000                 & 1000                       &        2345.99         &        383.96         &        $2^{(-)}$,$3^{(-)}$,$4^{(-)}$,$5^{(-)}$          &          965.88         &        135.52         &        $1^{(+)}$,$4^{(-)}$,$5^{(-)}$          &          995.26         &        165.62         &        $1^{(+)}$,$4^{(-)}$,$5^{(-)}$          &          130.83         &        22.98         &        $1^{(+)}$,$2^{(+)}$,$3^{(+)}$         &          130.05         &        38.08         &        $1^{(+)}$,$2^{(+)}$,$3^{(+)}$      \\
		& 100                        & 10000                 & 5000                       &        698.49         &        264.02         &        $4^{(-)}$,$5^{(-)}$          &          979.27         &        271.39         &        $4^{(-)}$,$5^{(-)}$          &          943.22         &        345.44         &        $4^{(-)}$,$5^{(-)}$          &          58.90         &        20.05         &        $1^{(+)}$,$2^{(+)}$,$3^{(+)}$         &          16.52         &        7.66         &        $1^{(+)}$,$2^{(+)}$,$3^{(+)}$      \\
		& 100                        & 10000                 & 15000                      &        299.81         &        156.20         &        $2^{(+)}$,$3^{(+)}$,$4^{(-)}$,$5^{(-)}$          &          977.11         &        467.85         &        $1^{(-)}$,$4^{(-)}$,$5^{(-)}$          &          956.92         &        659.17         &        $1^{(-)}$,$4^{(-)}$,$5^{(-)}$          &          32.70         &        20.88         &        $1^{(+)}$,$2^{(+)}$,$3^{(+)}$         &          5.21         &        4.16         &        $1^{(+)}$,$2^{(+)}$,$3^{(+)}$      \\ \hline
		unc-s-w & 100                        & 10000                 & 100                        &        5981.93         &        290.09         &        $2^{(-)}$,$3^{(-)}$,$4^{(-)}$,$5^{(-)}$          &          556.04         &        33.99         &        $1^{(+)}$,$4^{(-)}$,$5^{(-)}$          &          614.58         &        37.51         &        $1^{(+)}$,$4^{(-)}$,$5^{(-)}$          &          196.66         &        13.17         &        $1^{(+)}$,$2^{(+)}$,$3^{(+)}$         &          256.44         &        25.68         &        $1^{(+)}$,$2^{(+)}$,$3^{(+)}$      \\
		& 100                        & 10000                 & 1000                        &        998.81         &        250.14         &        $2^{(-)}$,$3^{(-)}$,$4^{(-)}$,$5^{(-)}$          &          631.46         &        109.27         &        $1^{(+)}$,$4^{(-)}$,$5^{(-)}$          &          591.64         &        109.36         &        $1^{(+)}$,$4^{(-)}$,$5^{(-)}$          &          81.59         &        20.33         &        $1^{(+)}$,$2^{(+)}$,$3^{(+)}$         &          54.06         &        16.60         &        $1^{(+)}$,$2^{(+)}$,$3^{(+)}$      \\
		& 100                        & 10000                 & 5000                       &        238.21         &        88.11         &        $2^{(+)}$,$3^{(+)}$,$4^{(-)}$,$5^{(-)}$          &          662.77         &        229.14         &        $1^{(-)}$,$4^{(-)}$,$5^{(-)}$          &          597.54         &        229.58         &        $1^{(-)}$,$4^{(-)}$,$5^{(-)}$          &          34.91         &        20.64         &        $1^{(+)}$,$2^{(+)}$,$3^{(+)}$         &          16.44         &        11.31         &        $1^{(+)}$,$2^{(+)}$,$3^{(+)}$      \\
		& 100                        & 10000                 & 15000                      &        69.81         &        33.58         &        $2^{(+)}$,$3^{(+)}$,$4^{(-)}$,$5^{(-)}$          &          640.56         &        292.14         &        $1^{(-)}$,$4^{(-)}$,$5^{(-)}$          &          569.05         &        285.58         &        $1^{(-)}$,$4^{(-)}$,$5^{(-)}$          &          10.84         &        10.10         &        $1^{(+)}$,$2^{(+)}$,$3^{(+)}$         &          3.91         &        5.01         &        $1^{(+)}$,$2^{(+)}$,$3^{(+)}$      \\ \hline
		bou-s-c & 100                        & 10000                 & 100                        &        9019.73         &        284.72         &        $2^{(-)}$,$4^{(-)}$,$5^{(-)}$          &          1456.79         &        37.26         &        $1^{(+)}$,$4^{(-)}$         &          1579.05         &        56.73         &        $4^{(-)}$,$5^{(-)}$          &          541.12         &        31.19         &        $1^{(+)}$,$2^{(+)}$,$3^{(+)}$         &          677.47         &        46.45         &        $1^{(+)}$,$3^{(+)}$      \\
		& 100                        & 10000                 & 1000                       &        2016.79         &        460.67         &        $2^{(-)}$,$4^{(-)}$,$5^{(-)}$          &          1451.67         &        121.58         &        $1^{(+)}$,$4^{(-)}$,$5^{(-)}$          &          1503.40         &        134.60         &        $4^{(-)}$,$5^{(-)}$          &          232.06         &        31.44         &        $1^{(+)}$,$2^{(+)}$,$3^{(+)}$         &          252.87         &        28.84         &        $1^{(+)}$,$2^{(+)}$,$3^{(+)}$      \\
		& 100                        & 10000        & 5000               &        786.47         &        308.14         &        $2^{(+)}$,$3^{(+)}$,$4^{(-)}$,$5^{(-)}$          &          1404.09         &        228.27         &        $1^{(-)}$,$4^{(-)}$,$5^{(-)}$          &          1489.08         &        269.61         &        $1^{(-)}$,$4^{(-)}$,$5^{(-)}$          &          124.10         &        36.70         &        $1^{(+)}$,$2^{(+)}$,$3^{(+)}$         &          126.24         &        29.41         &        $1^{(+)}$,$2^{(+)}$,$3^{(+)}$      \\
		& 100                        & 10000                 & 15000                      &        589.52         &        298.05         &        $2^{(+)}$,$3^{(+)}$,$4^{(-)}$,$5^{(-)}$          &          1346.87         &        309.54         &        $1^{(-)}$,$4^{(-)}$,$5^{(-)}$          &          1465.54         &        358.38         &        $1^{(-)}$,$4^{(-)}$,$5^{(-)}$          &          72.22         &        28.64         &        $1^{(+)}$,$2^{(+)}$,$3^{(+)}$         &          84.20         &        32.61         &        $1^{(+)}$,$2^{(+)}$,$3^{(+)}$                                 
	\end{tabular}
\end{sidewaystable}

\begin{sidewaystable}

	\setlength{\tabcolsep}{1pt}
	\centering
	\tiny
	
	\caption{The mean, standard deviation values and statistical tests of the partial offline error for \EAd, NSGA-II, SPEA2, NSGA-II with elitism, and SPEA2 with elitism based on the normal distribution.}
	\label{tbl:par_norm}
	\begin{tabular}{llllrrlrrlrrlrrlrrl}
		& \multicolumn{1}{c}{$n$} & \multicolumn{1}{c}{$\sigma$} & \multicolumn{1}{c}{$\tau$} & \multicolumn{3}{c}{\EAd (1)}     & \multicolumn{3}{c}{NSGA-II (2)}                                                 & \multicolumn{3}{c}{SPEA2 (3)}    & \multicolumn{3}{c}{NSGA-II(we) (4)}         & \multicolumn{3}{c}{SPEA2(we) (5)}                                                                \\
		&                            &                       &                            & \multicolumn{1}{c}{mean} & \multicolumn{1}{c}{st} & \multicolumn{1}{c}{stat} & \multicolumn{1}{c}{mean} & \multicolumn{1}{c}{st} & \multicolumn{1}{c}{stat} & \multicolumn{1}{c}{mean} & \multicolumn{1}{c}{st} & \multicolumn{1}{c}{stat} & \multicolumn{1}{c}{mean} & \multicolumn{1}{c}{st} & \multicolumn{1}{c}{stat} &
		\multicolumn{1}{c}{mean} &
		\multicolumn{1}{c}{st} & \multicolumn{1}{c}{stat}  \\ \hline
		uncor   & 100                   & 100                          & 100                       &        2850.69         &        502.97         &        $2^{(-)}$,$3^{(-)}$,$4^{(-)}$,$5^{(-)}$          &          7.43         &        1.75         &        $1^{(+)}$,$4^{(-)}$,$5^{(-)}$          &          5.59         &        2.25         &        $1^{(+)}$         &          4.99         &        2.14         &        $1^{(+)}$,$2^{(+)}$         &          4.36         &        2.76         &        $1^{(+)}$,$2^{(+)}$      \\
		& 100                   & 100                          & 1000                     &        190.16         &        22.90         &        $2^{(-)}$,$3^{(-)}$,$4^{(-)}$,$5^{(-)}$          &          3.41         &        2.00         &        $1^{(+)}$,$4^{(-)}$,$5^{(-)}$          &          1.47         &        1.24         &        $1^{(+)}$,$5^{(-)}$          &          0.45         &        0.22         &        $1^{(+)}$,$2^{(+)}$         &          0.25         &        0.16         &        $1^{(+)}$,$2^{(+)}$,$3^{(+)}$      \\
		& 100                   & 100                          & 5000                       &        45.93         &        12.35         &        $2^{(-)}$,$3^{(-)}$,$4^{(-)}$,$5^{(-)}$          &          4.73         &        3.54         &        $1^{(+)}$,$4^{(-)}$,$5^{(-)}$          &          2.46         &        2.54         &        $1^{(+)}$,$4^{(-)}$,$5^{(-)}$          &          0.11         &        0.13         &        $1^{(+)}$,$2^{(+)}$,$3^{(+)}$         &          0.00         &        0.00         &        $1^{(+)}$,$2^{(+)}$,$3^{(+)}$      \\
		& 100                   & 100                          & 15000                      &        26.30         &        12.86         &        $2^{(-)}$,$3^{(-)}$,$4^{(-)}$,$5^{(-)}$          &          4.19         &        3.49         &        $1^{(+)}$,$4^{(-)}$,$5^{(-)}$          &          1.80         &        2.39         &        $1^{(+)}$,$4^{(-)}$,$5^{(-)}$          &          0.06         &        0.11         &        $1^{(+)}$,$2^{(+)}$,$3^{(+)}$         &          0.00         &        0.00         &        $1^{(+)}$,$2^{(+)}$,$3^{(+)}$      \\ \hline
		unc-s-w & 100                   & 100                          & 100                        &        1837.47         &        854.78         &        $2^{(+)}$,$3^{(+)}$         &          7839.38         &        4620.58         &        $1^{(-)}$,$4^{(-)}$,$5^{(-)}$          &          7837.07         &        4647.17         &        $1^{(-)}$,$4^{(-)}$,$5^{(-)}$          &          3845.55         &        2425.39         &        $2^{(+)}$,$3^{(+)}$         &          3556.99         &        2159.38         &        $2^{(+)}$,$3^{(+)}$      \\
		& 100                   & 100                          & 1000                       &        1418.01         &        379.70         &        $2^{(+)}$,$3^{(+)}$,$4^{(+)}$,$5^{(+)}$          &          8892.21         &        1230.48         &        $1^{(-)}$,$4^{(-)}$,$5^{(-)}$          &          8980.88         &        1256.83         &        $1^{(-)}$,$4^{(-)}$,$5^{(-)}$          &          4243.08         &        988.88         &        $1^{(-)}$,$2^{(+)}$,$3^{(+)}$         &          4069.83         &        958.42         &        $1^{(-)}$,$2^{(+)}$,$3^{(+)}$      \\
		& 100                   & 100                          & 5000                       &        1258.84         &        413.82         &        $2^{(+)}$,$3^{(+)}$         &          4499.78         &        1488.58         &        $1^{(-)}$,$4^{(-)}$,$5^{(-)}$          &          4521.01         &        1498.07         &        $1^{(-)}$,$4^{(-)}$,$5^{(-)}$          &          2121.57         &        1216.61         &        $2^{(+)}$,$3^{(+)}$         &          2135.18         &        1218.82         &        $2^{(+)}$,$3^{(+)}$      \\
		& 100                   & 100                          & 15000                      &        1083.84         &        649.82         &        $2^{(+)}$,$3^{(+)}$         &          4698.90         &        1168.47         &        $1^{(-)}$,$4^{(-)}$,$5^{(-)}$          &          4705.49         &        1163.00         &        $1^{(-)}$,$4^{(-)}$,$5^{(-)}$          &          1888.21         &        1330.09         &        $2^{(+)}$,$3^{(+)}$         &          2014.72         &        1485.71         &        $2^{(+)}$,$3^{(+)}$      \\ \hline
		bou-s-c & 100                   & 100                          & 100                        &        1393.46         &        183.93         &        $2^{(-)}$,$3^{(-)}$,$4^{(-)}$,$5^{(-)}$          &          18.70         &        2.02         &        $1^{(+)}$,$4^{(-)}$,$5^{(-)}$          &          14.18         &        1.65         &        $1^{(+)}$,$4^{(-)}$,$5^{(-)}$          &          8.20         &        1.87         &        $1^{(+)}$,$2^{(+)}$,$3^{(+)}$         &          7.33         &        1.69         &        $1^{(+)}$,$2^{(+)}$,$3^{(+)}$      \\
		& 100                   & 100                          & 1000                       &        138.05         &        28.21         &        $3^{(-)}$,$4^{(-)}$,$5^{(-)}$          &          17.13         &        0.85         &        $4^{(-)}$,$5^{(-)}$          &          12.47         &        0.98         &        $1^{(+)}$,$4^{(-)}$,$5^{(-)}$          &          2.20         &        0.63         &        $1^{(+)}$,$2^{(+)}$,$3^{(+)}$         &          2.10         &        1.09         &        $1^{(+)}$,$2^{(+)}$,$3^{(+)}$      \\
		& 100                   & 100                          & 5000                      &        52.09         &        14.34         &        $3^{(-)}$,$4^{(-)}$,$5^{(-)}$          &          17.15         &        1.23         &        $4^{(-)}$,$5^{(-)}$          &          12.74         &        2.07         &        $1^{(+)}$,$4^{(-)}$,$5^{(-)}$          &          0.67         &        0.30         &        $1^{(+)}$,$2^{(+)}$,$3^{(+)}$         &          0.38         &        0.29         &        $1^{(+)}$,$2^{(+)}$,$3^{(+)}$      \\
		& 100                   & 100                          & 15000                      &        41.86         &        20.77         &        $3^{(-)}$,$4^{(-)}$,$5^{(-)}$          &          17.39         &        2.04         &        $4^{(-)}$,$5^{(-)}$          &          13.19         &        2.20         &        $1^{(+)}$,$4^{(-)}$,$5^{(-)}$          &          0.35         &        0.23         &        $1^{(+)}$,$2^{(+)}$,$3^{(+)}$         &          0.06         &        0.07         &        $1^{(+)}$,$2^{(+)}$,$3^{(+)}$      \\ \hline
		uncor   & 100                   & 500                          & 100                        &        1350.69         &        422.15         &        $2^{(-)}$,$3^{(-)}$,$4^{(-)}$,$5^{(-)}$          &          54.77         &        29.73         &        $1^{(+)}$,$4^{(-)}$,$5^{(-)}$          &          40.24         &        23.67         &        $1^{(+)}$,$4^{(-)}$,$5^{(-)}$          &          7.22         &        3.63         &        $1^{(+)}$,$2^{(+)}$,$3^{(+)}$         &          5.70         &        3.66         &        $1^{(+)}$,$2^{(+)}$,$3^{(+)}$      \\
		& 100                   & 500                          & 1000                       &        183.02         &        88.65         &        $3^{(-)}$,$4^{(-)}$,$5^{(-)}$          &          70.71         &        52.99         &        $4^{(-)}$,$5^{(-)}$          &          51.71         &        39.88         &        $1^{(+)}$,$4^{(-)}$,$5^{(-)}$          &          0.70         &        0.52         &        $1^{(+)}$,$2^{(+)}$,$3^{(+)}$         &          0.20         &        0.12         &        $1^{(+)}$,$2^{(+)}$,$3^{(+)}$      \\
		& 100                   & 500                          & 5000                       &        23.90         &        14.52         &        $2^{(+)}$,$3^{(+)}$,$4^{(-)}$,$5^{(-)}$          &          108.64         &        59.69         &        $1^{(-)}$,$4^{(-)}$,$5^{(-)}$          &          81.33         &        47.77         &        $1^{(-)}$,$4^{(-)}$,$5^{(-)}$          &          0.12         &        0.11         &        $1^{(+)}$,$2^{(+)}$,$3^{(+)}$         &          0.01         &        0.02         &        $1^{(+)}$,$2^{(+)}$,$3^{(+)}$      \\
		& 100                   & 500                          & 15000                      "&        1.59         &        1.27         &        $2^{(+)}$,$3^{(+)}$,$4^{(-)}$,$5^{(-)}$          &          144.02         &        64.91         &        $1^{(-)}$,$4^{(-)}$,$5^{(-)}$          &          113.76         &        51.80         &        $1^{(-)}$,$4^{(-)}$,$5^{(-)}$          &          0.04         &        0.13         &        $1^{(+)}$,$2^{(+)}$,$3^{(+)}$         &          0.00         &        0.00         &        $1^{(+)}$,$2^{(+)}$,$3^{(+)}$      \\ \hline
		unc-s-w & 100                   & 500                          & 100                        &        1884.04         &        513.97         &        $2^{(-)}$,$3^{(-)}$,$4^{(-)}$,$5^{(-)}$          &          18.76         &        6.17         &        $1^{(+)}$,$3^{(+)}$,$5^{(+)}$          &          159.74         &        61.22         &        $1^{(+)}$,$2^{(-)}$,$4^{(-)}$         &          17.92         &        5.98         &        $1^{(+)}$,$3^{(+)}$,$5^{(+)}$          &          71.71         &        27.52         &        $1^{(+)}$,$2^{(-)}$,$4^{(-)}$      \\
		& 100                   & 500                          & 1000                       &        165.71         &        110.24         &        $2^{(-)}$,$4^{(-)}$         &          2.62         &        1.14         &        $1^{(+)}$,$3^{(+)}$,$5^{(+)}$          &          428.28         &        231.68         &        $2^{(-)}$,$4^{(-)}$,$5^{(-)}$          &          1.39         &        0.84         &        $1^{(+)}$,$3^{(+)}$,$5^{(+)}$          &          45.54         &        20.00         &        $2^{(-)}$,$3^{(+)}$,$4^{(-)}$      \\
		& 100                   & 500                          & 5000                       &        13.86         &        10.22         &        $2^{(-)}$,$3^{(+)}$,$4^{(-)}$         &          1.70         &        1.11         &        $1^{(+)}$,$3^{(+)}$,$4^{(-)}$,$5^{(+)}$          &          424.22         &        348.57         &        $1^{(-)}$,$2^{(-)}$,$4^{(-)}$,$5^{(-)}$          &          0.06         &        0.10         &        $1^{(+)}$,$2^{(+)}$,$3^{(+)}$,$5^{(+)}$          &          14.61         &        13.33         &        $2^{(-)}$,$3^{(+)}$,$4^{(-)}$      \\
		& 100                   & 500                          & 15000                      &        2.40         &        2.04         &        $3^{(+)}$,$4^{(-)}$         &          2.22         &        2.19         &        $3^{(+)}$,$4^{(-)}$         &          280.57         &        302.69         &        $1^{(-)}$,$2^{(-)}$,$4^{(-)}$,$5^{(-)}$          &          0.02         &        0.08         &        $1^{(+)}$,$2^{(+)}$,$3^{(+)}$,$5^{(+)}$          &          3.17         &        4.74         &        $3^{(+)}$,$4^{(-)}$      \\\hline
		bou-s-c & 100                   & 500                          & 100                        &        907.16         &        165.35         &        $3^{(-)}$,$4^{(-)}$,$5^{(-)}$          &          100.75         &        5.39         &        $4^{(-)}$,$5^{(-)}$          &          80.71         &        3.23         &        $1^{(+)}$,$5^{(-)}$          &          19.01         &        1.55         &        $1^{(+)}$,$2^{(+)}$         &          13.05         &        2.21         &        $1^{(+)}$,$2^{(+)}$,$3^{(+)}$      \\
		& 100                   & 500                          & 1000                       &        129.70         &        36.54         &        $3^{(-)}$,$4^{(-)}$,$5^{(-)}$          &          106.58         &        10.22         &        $3^{(-)}$,$4^{(-)}$,$5^{(-)}$          &          81.95         &        6.81         &        $1^{(+)}$,$2^{(+)}$,$4^{(-)}$,$5^{(-)}$          &          4.32         &        1.44         &        $1^{(+)}$,$2^{(+)}$,$3^{(+)}$         &          1.70         &        1.32         &        $1^{(+)}$,$2^{(+)}$,$3^{(+)}$      \\
		& 100                   & 500                          & 5000                       &        31.63         &        20.82         &        $2^{(+)}$,$4^{(-)}$,$5^{(-)}$          &          113.69         &        15.02         &        $1^{(-)}$,$4^{(-)}$,$5^{(-)}$          &          85.27         &        9.19         &        $4^{(-)}$,$5^{(-)}$          &          0.86         &        0.39         &        $1^{(+)}$,$2^{(+)}$,$3^{(+)}$         &          0.10         &        0.12         &        $1^{(+)}$,$2^{(+)}$,$3^{(+)}$      \\
		& 100                   & 500                          & 15000                      &        7.40         &        3.70         &        $2^{(+)}$,$3^{(+)}$,$5^{(-)}$          &          119.19         &        14.98         &        $1^{(-)}$,$4^{(-)}$,$5^{(-)}$          &          89.74         &        13.50         &        $1^{(-)}$,$4^{(-)}$,$5^{(-)}$          &          0.33         &        0.28         &        $2^{(+)}$,$3^{(+)}$         &          0.01         &        0.01         &        $1^{(+)}$,$2^{(+)}$,$3^{(+)}$                                 
	\end{tabular}

\end{sidewaystable}

In Table \ref{tbl:tot_norm}, we compare these algorithms for different instances based on the normal distribution. We investigate the combinations; $\sigma = 100$ and $\sigma = 500$, and $\tau = 100, 1000, 50000, 150000$. \EAd is usually outperformed by SPEA2. However, with the exception of the following cases: for uncorrelated similar weights with $\sigma =100$, $\tau = 5000, 1500$, and for uncorrelated similar weights with $\sigma =500$, $\tau = 5000, 1500$, \EAd has performed better. Furthermore, by comparing \EAd and NSGA-II, we observe the same behavior except for the mean values for the instance bounded strongly correlated where $ \sigma = 500$, $\tau = 1500$ are substantially worse than those of \EAd. The results show that NSGA-II(we) and SPEA2(we) significantly outperform \EAd, NSGA-II and SPEA2 in most of the cases as expected due to the elitism approach.

We now considering the new partial offline performance measure. Table \ref{tbl:par_unif} clearly indicates that the elitism mechanisms used in NSGA-II(we) and SPEA2(we) achieve substantially better results compared NSGA-II, SPEA2, and \EAd in all instances. Interestingly, NSGA-II(we) benefits much more from elitism than SPEA2(we), and NSGA-II(we) achieves the lowest partial offline error across most instances. An exception are bounded strongly correlated instances with $r = 100$. \EAd, especially for $r = 2000, 10000$ and $\tau = 15000$ significantly outperforms NSGA-II and SPEA2 by one order of magnitude in terms of offline error in most cases.

Furthermore, Table \ref{tbl:par_norm} shows results for the normal distribution. The results summarize the partial offline error and statistical tests for all five algorithms. NSGA-II(we) and SPEA2(we) perform significantly better than \EAd, NSGA-II and SPEA2 in all instances. Noticeably, SPEA2(we) in following instances; uncorrelated and bounded strongly correlated, and $\sigma =500$, $\tau = 5000, 15000$ finds the results close to the optimum. Similar to the experiments based on the total offline error, \EAd is outperformed by NSGA-II and SPEA2, whereas \EAd outperforms NSGA-II and SPEA2 in the instance with uncorrelated similar weights and $\sigma =100$.

The impact of elitism in \EAd, NSGA-II(we), and SPEA2(we) is also illustrated by observing the mean values of the same column in one specific instance. Note that in an environment with low frequency changes the algorithms have more time to find a good solution before the next change happens, i.e., the mean of total/partial offline error for one specific algorithm against the same instance supposed to decrease by increasing the value of $\tau$. However, such pattern could not be seen for NSGA-II and SPEA2, mostly in $\tau\geq1000$. It shows that NSGA-II and SPEA2 find and fix a well distributed set of solution in the first $1000$ iterations, according to the classic distribution techniques, and do not improve the best found solution, which cause adding the same amount of error to the total offline error in the next iterations.

Comparing the results of algorithms in uniform instances according the total and offline partial errors (Tables \ref{tbl:tot_unif},\ref{tbl:par_unif}), also illustrates the the importance of customized elitism in advanced evolutionary algorithm. The outperformance of \EAd in low frequency changes according to the partial offline error shows that at the end of each interval, it has found a better solution than classic NSGA-II and SPEA2. However, according to the total offline error, this progress is slow and not significantly better in average.

Overall, our results suggest that NSGA-II(we) and SPEA2(we) using the elitism mechanism significantly outperform the classical NSGA-II, SPEA2 and \EAd. Furthermore, our experiments confirmed that in environments with medium and high frequency changes \EAd is outperformed by NSGA-II and SPEA2. However, there is no significant difference between the performance of \EAd and classic versions of NSGA-II and SPEA2 when there is sufficient time for algorithms to adapt their solutions according to the new change.

\section{Conclusions and Future Work}\label{sec:conc}
In this paper we studied the evolutionary algorithms for the KP where the capacity dynamically changes during the optimization process. In the introduced dynamic setting, the frequency of changes is determined by $\tau$. The magnitude of changes is chosen randomly either under the uniform distribution $\mathcal{U}(-r, r)$ or under the normal distribution $\mathcal{N}(0, \sigma^2)$. We compared the performance of (1+1)~EA, two simple multi-objective approaches with different dominance definitions (\EA, \EAd), the classic versions of NSGA-II and SPEA2 as advanced multi-objective algorithms, and the effect of customized elitism on their performance. Our experiments in the case of uniform weights verified the previous theoretical studies for (1+1)~EA and \EA~\cite{DBLP:journals/algorithmica/ShiSFKN19}. It is shown that the multi-objective approach, which uses a population in the optimization, outperforms (1+1)~EA. Furthermore, we considered the algorithms in the case of general weights for different classes of instances with a variation of frequencies and magnitudes. Our results illustrated that \EA does not perform well in the general case due to its dominance procedure. However, \EAd, which benefits from a population with a smaller size and non-dominated solutions, beats (1+1)~EA in most cases. On the other hand, in the environments with highly frequency changes, (1+1)~EA performs better than the multi-objective approaches. In such cases, the population slows down \EAd in reacting to the dynamic change. Selecting \EAd as the winner baseline algorithm, we compared its performance with NSGA-II and SPEA2 as advanced evolutionary algorithm. Our results showed that although classic versions of NSGA-II and SPEA2 are significantly better than \EAd in many cases, their distribution handling techniques prevent the algorithms from keep track of the optimal solutions. Thus, \EAd outperform them in low frequency changes. To address this weakness, we improved these algorithms by applying an additional elitism which keeps the best found solution in the population. The final results show that SPEA2 and NSGA-II with new elitism approach outperform \EAd even facing high frequency changes.

\section*{Acknowledgment}
This work has been supported by the Australian Research Council (ARC) through grant DP160102401 and by the South Australian Government through the Research Consortium "Unlocking Complex Resources through Lean Processing".

\bibliographystyle{unsrt}
\bibliography{ecjsample.bib}

\begin{thebibliography}{10}

\bibitem{DBLP:series/ncs/EibenS15}
A.~E. Eiben and James~E. Smith.
\newblock {\em Introduction to Evolutionary Computing, Second Edition}.
\newblock Natural Computing Series. Springer, 2015.

\bibitem{DCOPS}
T.T. Nguyen and X.~Yao.
\newblock Continuous dynamic constrained optimization: The challenges.
\newblock {\em IEEE Transactions on Evolutionary Computation}, 16(6):769--786,
  2012.

\bibitem{DBLP:journals/swevo/RakshitKD17}
Pratyusha Rakshit, Amit Konar, and Swagatam Das.
\newblock Noisy evolutionary optimization algorithms - {A} comprehensive
  survey.
\newblock {\em Swarm and Evolutionary Computation}, 33:18--45, 2017.

\bibitem{DBLP:conf/aaai/DoerrD0NS20}
Benjamin Doerr, Carola Doerr, Aneta Neumann, Frank Neumann, and Andrew~M.
  Sutton.
\newblock Optimization of chance-constrained submodular functions.
\newblock In {\em The Thirty-Fourth {AAAI} Conference on Artificial
  Intelligence, {AAAI} 2020}, pages 1460--1467. {AAAI} Press, 2020.

\bibitem{DBLP:conf/ppsn/NeumannN20}
Aneta Neumann and Frank Neumann.
\newblock Optimising monotone chance-constrained submodular functions using
  evolutionary multi-objective algorithms.
\newblock In {\em Parallel Problem Solving from Nature - {PPSN} {XVI},
  Proceedings, Part {I}}, volume 12269 of {\em LNCS}, pages 404--417. Springer,
  2020.

\bibitem{DBLP:journals/swevo/NguyenYB12}
Trung~Thanh Nguyen, Shengxiang Yang, and J{\"{u}}rgen Branke.
\newblock Evolutionary dynamic optimization: {A} survey of the state of the
  art.
\newblock {\em Swarm and Evolutionary Computation}, 6:1--24, 2012.

\bibitem{DBLP:conf/evoW/Ameca-AlducinHN18}
Maria~Yaneli Ameca{-}Alducin, Maryam Hasani{-}Shoreh, and Frank Neumann.
\newblock On the use of repair methods in differential evolution for dynamic
  constrained optimization.
\newblock In {\em Applications of Evolutionary Computation, Proceedings},
  volume 10784 of {\em LNCS}, pages 832--847. Springer, 2018.

\bibitem{DBLP:journals/corr/abs-1806-08547}
Vahid Roostapour, Mojgan Pourhassan, and Frank Neumann.
\newblock Analysis of evolutionary algorithms in dynamic and stochastic
  environments.
\newblock {\em CoRR}, abs/1806.08547, 2018.

\bibitem{DBLP:conf/gecco/PourhassanGN15}
Mojgan Pourhassan, Wanru Gao, and Frank Neumann.
\newblock Maintaining 2-approximations for the dynamic vertex cover problem
  using evolutionary algorithms.
\newblock In {\em Proceedings of the Genetic and Evolutionary Computation
  Conference, {GECCO} 2015}, pages 903--910. {ACM}, 2015.

\bibitem{DBLP:conf/gecco/Bossek0PS19}
Jakob Bossek, Frank Neumann, Pan Peng, and Dirk Sudholt.
\newblock Runtime analysis of randomized search heuristics for dynamic graph
  coloring.
\newblock In {\em Proceedings of the Genetic and Evolutionary Computation
  Conference, {GECCO} 2019}, pages 1443--1451. {ACM}, 2019.

\bibitem{DBLP:conf/aaai/RoostapourN0019}
Vahid Roostapour, Aneta Neumann, Frank Neumann, and Tobias Friedrich.
\newblock Pareto optimization for subset selection with dynamic cost
  constraints.
\newblock In {\em The Thirty-Third {AAAI} Conference on Artificial
  Intelligence, {AAAI} 2019}, pages 2354--2361. {AAAI} Press, 2019.

\bibitem{DBLP:journals/algorithmica/ShiSFKN19}
Feng Shi, Martin Schirneck, Tobias Friedrich, Timo K{\"{o}}tzing, and Frank
  Neumann.
\newblock Reoptimization time analysis of evolutionary algorithms on linear
  functions under dynamic uniform constraints.
\newblock {\em Algorithmica}, 81(2):828--857, 2019.

\bibitem{DBLP:conf/gecco/BrankeSU05}
J{\"{u}}rgen Branke, Erdem Salihoglu, and Sima Uyar.
\newblock Towards an analysis of dynamic environments.
\newblock In {\em Proceedings of the Genetic and Evolutionary Computation
  Conference, {GECCO} 2005}, pages 1433--1440. {ACM}, 2005.

\bibitem{10.1007/11732242_74}
J{\"u}rgen Branke, Merve Orbay{\i}, and {\c{S}}ima Uyar.
\newblock The role of representations in dynamic knapsack problems.
\newblock In {\em Applications of Evolutionary Computing}, pages 764--775.
  Springer Berlin Heidelberg, 2006.

\bibitem{10.1007/978-3-642-01129-0_86}
{\c{S}}ima Uyar and H.~Turgut Uyar.
\newblock A critical look at dynamic multi-dimensional knapsack problem
  generation.
\newblock In {\em Applications of Evolutionary Computing}, pages 762--767.
  Springer Berlin Heidelberg, 2009.

\bibitem{DBLP:journals/nc/NeumannW06}
Frank Neumann and Ingo Wegener.
\newblock Minimum spanning trees made easier via multi-objective optimization.
\newblock {\em Nat. Comput.}, 5(3):305--319, 2006.

\bibitem{DBLP:journals/ec/FriedrichHHNW10}
Tobias Friedrich, Jun He, Nils Hebbinghaus, Frank Neumann, and Carsten Witt.
\newblock Approximating covering problems by randomized search heuristics using
  multi-objective models.
\newblock {\em Evol. Comput.}, 18(4):617--633, 2010.

\bibitem{DBLP:journals/algorithmica/KratschN13}
Stefan Kratsch and Frank Neumann.
\newblock Fixed-parameter evolutionary algorithms and the vertex cover problem.
\newblock {\em Algorithmica}, 65(4):754--771, 2013.

\bibitem{DBLP:journals/ec/FriedrichN15}
Tobias Friedrich and Frank Neumann.
\newblock Maximizing submodular functions under matroid constraints by
  evolutionary algorithms.
\newblock {\em Evol. Comput.}, 23(4):543--558, 2015.

\bibitem{DBLP:books/sp/ZhouYQ19}
Zhi{-}Hua Zhou, Yang Yu, and Chao Qian.
\newblock {\em Evolutionary Learning: Advances in Theories and Algorithms}.
\newblock Springer, 2019.

\bibitem{DBLP:conf/ppsn/Roostapour0N18}
Vahid Roostapour, Aneta Neumann, and Frank Neumann.
\newblock On the performance of baseline evolutionary algorithms on the dynamic
  knapsack problem.
\newblock In {\em Parallel Problem Solving from Nature - {PPSN} {XV},
  Proceedings, Part {I}}, volume 11101 of {\em LNCS}, pages 158--169. Springer,
  2018.

\bibitem{DBLP:journals/tec/DebAPM02}
Kalyanmoy Deb, Samir Agrawal, Amrit Pratap, and T.~Meyarivan.
\newblock A fast and elitist multiobjective genetic algorithm: {NSGA-II}.
\newblock {\em {IEEE} Trans. Evolutionary Computation}, 6(2):182--197, 2002.

\bibitem{zitzler2001spea2}
Eckart Zitzler, Marco Laumanns, and Lothar Thiele.
\newblock {SPEA2}: Improving the strength pareto evolutionary algorithm.
\newblock {\em TIK-report,}, 103:1--22, 2001.

\bibitem{DBLP:journals/aes/DurilloN11}
Juan~Jos{\'{e}} Durillo and Antonio~J. Nebro.
\newblock j{M}etal: {A} java framework for multi-objective optimization.
\newblock {\em Advances in Engineering Software}, 42(10):760--771, 2011.

\bibitem{polyakovskiy2014comprehensive}
Sergey Polyakovskiy, Mohammad~Reza Bonyadi, Markus Wagner, Zbigniew
  Michalewicz, and Frank Neumann.
\newblock A comprehensive benchmark set and heuristics for the traveling thief
  problem.
\newblock In {\em Proceedings of Conference on Genetic and Evolutionary
  Computation}, pages 477--484. ACM, 2014.

\bibitem{Corder09}
Gregory~W. Corder and Dale~I. Foreman.
\newblock {\em {Nonparametric Statistics for Non-Statisticians: A Step-by-Step
  Approach}}.
\newblock Wiley, 2009.

\bibitem{DBLP:journals/algorithmica/ShiSFKN20}
Feng Shi, Martin Schirneck, Tobias Friedrich, Timo K{\"{o}}tzing, and Frank
  Neumann.
\newblock Correction to: Reoptimization time analysis of evolutionary
  algorithms on linear functions under dynamic uniform constraints.
\newblock {\em Algorithmica}, 82(10):3117--3123, 2020.

\bibitem{DBLP:books/daglib/0025643}
Frank Neumann and Carsten Witt.
\newblock {\em Bioinspired Computation in Combinatorial Optimization}.
\newblock Natural Computing Series. Springer, 2010.

\end{thebibliography}

\end{document}